\documentclass{{sig-alternate-05-2015}}
\usepackage{mathtools,xparse}
\usepackage{graphicx,dblfloatfix}     %

\newcounter{tisctr}
\renewcommand{\thetisctr}{\arabic{tisctr}}

\usepackage{array}
\newcolumntype{H}{>{\setbox0=\hbox\bgroup}c<{\egroup}@{}}

\DeclarePairedDelimiterX{\norm}[1]{\lvert}{\rvert}{#1}
\DeclarePairedDelimiterX{\normF}[1]{\lVert}{\rVert}{#1}

\newcommand{\algo}{\emph{PFR} }

\newcommand{\spara}[1]{\smallskip\noindent\textbf{#1}}

	{\normalsize \end{list} 
		}

\newenvironment{defn-eqn}[2]{
	\vskip 8pt 
	\refstepcounter{tisctr} %
	\noindent {\textbf{Definition \thetisctr.~}(#1) \emph{#2}} \vspace{2mm}}
	
\newenvironment{defn-test}[2]{\vskip 6pt \refstepcounter{tisctr} %
	\noindent {{\bf Definition \thetisctr.~}(#1) \emph{#2}}%
	\vskip 0pt
	\begin{equation}{}{\labelwidth 0pt \labelsep 0pt%
			\parsep 0pt
		}}%
		{\normalsize \end{equation} 
		}

\newcommand{\squishlist}{
	\begin{list}{$\bullet$}
		{ \setlength{\itemsep}{0pt}
			\setlength{\parsep}{1pt}
			\setlength{\topsep}{1pt}
			\setlength{\partopsep}{0pt}
			\setlength{\leftmargin}{1.5em}
			\setlength{\labelwidth}{1em}
			\setlength{\labelsep}{0.5em} } }
	\newcommand{\squishend}{\end{list}}

 \usepackage{amsmath}
\usepackage{xstring}
\usepackage[show]{notes}
\usepackage{booktabs}
\usepackage[normalem]{ulem}
\useunder{\uline}{\ul}{}
\usepackage{hyperref}
\usepackage{adjustbox}
\usepackage{subcaption}
\usepackage{colortbl}
\graphicspath{{./plots/}}
\usepackage{balance}
\usepackage{comment}
\usepackage[labelfont=bf]{caption}

\begin{document}
\title{Operationalizing Individual Fairness \\ with Pairwise Fair Representations}	
	
\numberofauthors{3}
\author{
	\alignauthor
		Preethi Lahoti\\
		\affaddr{Max Planck Institute for Informatics \\
			Saarland Informatics Campus\\}
		\affaddr{Saarbr\"{u}cken, Germany}\\
		\email{plahoti@mpi-inf.mpg.de}
	\alignauthor
	Krishna P. Gummadi\\
		\affaddr{Max Planck Institute for Software Systems \\
			Saarland Informatics Campus\\}
		\affaddr{Saarbr\"{u}cken, Germany}
		\email{gummadi@mpi-sws.org}
	\and
	\alignauthor
	Gerhard Weikum\\
		\affaddr{Max Planck Institute for Informatics \\
			Saarland Informatics Campus\\}
		\affaddr{Saarbr\"{u}cken, Germany}\\
		\email{weikum@mpi-inf.mpg.de}
	}
	
\maketitle	

\begin{abstract}
We revisit the notion of individual fairness proposed by 
Dwork et al.
A central challenge in operationalizing their approach is the difficulty in eliciting a human specification of a {similarity metric}. 
In this paper, we propose an operationalization of individual fairness that does not rely on a human specification of a distance metric. 
Instead, we propose novel approaches to elicit and leverage side-information on 
equally deserving individuals to counter subordination between social groups.
We model this knowledge as a {fairness graph}, and learn a unified
{Pairwise Fair Representation} (PFR) of the data that captures both data-driven similarity between individuals and the pairwise side-information in fairness graph.
We elicit fairness judgments from a variety of sources, including human judgments for two real-world datasets on recidivism prediction (COMPAS) and violent neighborhood prediction (Crime \& Communities). 
Our experiments show that 
the PFR model
for operationalizing
individual fairness 
is practically viable.
\footnote{This is a preprint of a full paper in the proceedings of the VLDB Endowment, Vol. 13, No. 4.}

\end{abstract}

\section{Introduction}

\subsection{Motivation} \label{sec-motivation}

Machine learning based prediction and ranking models are playing an
increasing role in decision making scenarios that affect human lives.
Examples include loan approval decisions in banking, candidate rankings
in employment, welfare benefit determination in social services, and
recidivism risk prediction
in criminal justice.
The societal impact of these algorithmic decisions has raised
concerns about their
fairness \cite{angwin2016machine,crawford2016artificial}, and 
recent research has started to investigate how to incorporate
formalized notions of fairness into machine prediction models
(e.g., \cite{dwork2012fairness,hardt2016equality,kamishima2012considerations,zafar2017fairness}).

\spara{\bf Individual vs Group Fairness:}  The fairness
notions explored by the bulk of the works can be broadly categorized
as targeting either {\em group fairness}~\cite{pedreshi2008discrimination,feldman2015certifying} 
or {\em individual
	fairness}~\cite{dwork2012fairness}. Group fairness notions attempt to ensure that
members of all protected groups in the population (e.g., based on
demographic attributes like gender or race) receive their ``fair share
of beneficial outcomes'' in a downstream task. 
To this end, one or more {\em protected attributes} and respective values
are specified, and given special treatment in machine learning models.
Numerous operationalizations of group fairness have been
proposed and evaluated including demographic parity~\cite{feldman2015certifying}, equality
of opportunity~\cite{hardt2016equality}, equalized odds~\cite{hardt2016equality}, and envy-free group
fairness~\cite{zafar2017parity}. These operationalizations differ in the measures used
to quantify a group's ``fair share of beneficial outcomes'' as well as
the mechanisms used to optimize for the fairness measures.

While effective at countering group-based discrimination in
decision outcomes, 
group fairness notions do not address unfairness in
outcomes at the level of individual users. For instance, it is natural
for individuals to compare their outcomes with those of others with
similar qualifications (independently of their group membership) and
perceive any differences in outcomes amongst individuals with similar
standing as unfair. 
\spara{Individual Fairness:} In their seminal work \cite{dwork2012fairness}, Dwork et al.  introduced a powerful notion of fairness called individual fairness, which states that ``similar individuals should be treated similarly''. In the original form of individual fairness introduced in \cite{dwork2012fairness}, the authors envisioned that a task-specific similarity metric would be provided by human experts which captures the similarity between individuals 
(e.g., ``a student who studies at University W and has a GPA X is
similar to another student who studies at University Y and has GPA
Z'').  
The individual fairness notion stipulates that individuals who
are deemed {similar} according to this \emph{task-specific similarity
  metric} should receive similar outcomes. Operationalizing this
strong notion of fairness can help in avoiding unfairness at an
individual level.

However, eliciting such a quantitative measure of similarity 
from humans has been the most challenging aspect of the individual
fairness framework, and little progress has been made on this open
problem. Two noteworthy subsequent works on individual fairness are
\cite{zemel2013learning} and \cite{lahoti2018ifair},
wherein the authors operationalize a simplified notion of similarity
metric. Concretely, they assume a distance metric (similarity metric)
such as a \emph{weighted} Euclidean distance over a feature space of
data atttributes, and aim to learn \emph{fair feature weights} for
this distance metric. This simplification of the individual fairness
notion largely limits the scope of the original idea of
\cite{dwork2012fairness}: ``\dots a (near ground-truth) approximation agreed upon by the society of the extent to which two individuals are deemed similar with respect to the task \dots''.
In this work we revisit the original notion of individual fairness. There are two main challenges in its operationalization: 
First, it is very difficult, if not impossible for humans to come up with a precise quantitative similarity metric that can be used to measure ``who is similar to whom''. 
Second, even if we assume that humans are capable of giving a precise similarity metric, it is still challenging for experts to model subjective side-information such as ``who should be treated similar to whom'' 
as a 
quantitative
similarity metric. 

\spara{\bf Examples:}
The challenge is illustrated by two scenarios:
\begin{itemize}
\item Consider the task of selecting researchers for academic jobs. Due to the difference in
publication culture of various communities, 
the citation counts of \emph{successful} researchers in programming language are known to be typically lower than that of \emph{successful} machine learning researchers. An expert 
recruiter
might have the background information for fair selection that ``an ML researcher with high citations is 
similarly strong and thus equally deserving as a
PL researcher with relatively lower citations''. It is 
all but 
easy to specify this background knowledge 
as a 
similarity metric. 
\item Consider the task of selecting students for Graduate School in the US. 
It is well known that SAT tests can be taken multiple times, and only the best score is reported for admissions. 
Further, each attempt to re-take the SAT test comes at a financial cost.
Due to complex interplay of historical subordination and social circumstances, it is known that, on average, SAT scores for 
African-American students are lower than for white students \cite{brooks1992rethinking}. 
Keeping anti-subordination in mind, a fairness expert might deem an African-American student with a relatively lower SAT score to be similar to and equally deserving as a white student with a slightly higher score. Once again, it is not easy to model this information as a 
similarity 
metric.
\end{itemize}

\spara{Research Questions:} We address 
the following research questions in this paper.
\squishlist
\item[-] [RQ1] How to elicit and model 
various kinds of 
background information on individual fairness?

\item[-] [RQ2] How to encode this background information, such that downstream tasks can make use of it for data-driven predictions and decision making?
\squishend

\subsection{Approach}
\spara{[RQ1] From Distance Metric to Fairness Graph.} 

\spara{Key Idea:} It is difficult, if not impossible, for human experts to judge ``the extent to which two individuals are similar'', much less formulate a precise \emph{similarity metric}.
In this paper, we posit that it is much easier for experts to make pairwise judgments about who 
is equally deserving and 
should be treated similar to whom. 

We propose to capture these pairwise judgments as a {\it fairness graph}, $G$, with edges between pairs of
individuals deemed similar with respect to the given task. 
{\color{black}We view this as valuable side information, but we consider it to be subjective and noisy.
Aggregation over many users can mitigate this, but we cannot expect $G$ to be perfectly fair.
Further, for generality, we do not assume that these are always complete.
In many applications, only partial and sometimes sparse 
fairness judgments would be available.}
In our experiments, we study the sensitivity to the amount of data in $G$
in Subsection \ref{subsec:sensitivity_to_labelsize}.
In Section \ref{sec:fairness_graphs} we address some of the practical challenges that arise in eliciting pairwise judgments such as comparing 
individuals from diverse 
groups, and we present various methods to construct fairness graphs. 

It is worth highlighting that we only need pairwise judgments for a small sample of individuals in the training data for the application task.
Naturally, no human judgments are elicited for test data (unseen data).
So once the prediction model for the application at hand has been learned,
only the regular data attributes of individuals are needed.

\spara{[RQ2] Learning Pairwise Fair Representations.} 

Given a fairness graph $G$, the goal of an individually fair
algorithm is to minimize the inconsistency (differences) in
outcomes for pairs of individuals connected in graph $G$. Thus, every
edge in graph $G$ represents a fairness constraint that the
algorithm needs to satisfy.
In Section \ref{sec-model}, we propose a model called \algo (for Pairwise Fair Representations),
which learns a new data representation with the aim of preserving the
utility of the input feature space (i.e., retaining as much
information of the input as possible), while incorporating the
 fairness 
 constraints captured 
 in the fairness graph.

Specifically, \algo aims to learn a latent data representation that preserves the local neighborhoods in the input data space,
while ensuring that individuals connected in the fairness graph are mapped to nearby points in the learned representation.
Since local neighborhoods in the learned representation capture individual fairness, once a fair representation is learned, any out-of-the-box downstream predictor can be directly applied.
\algo
takes as input
\squishlist
\item data records for individuals in the form of a feature matrix $X$ for training
a predictor, and
\item a (sparse) fairness graph $G$ that captures pairwise similarity for a 
subsample
of individuals in the training data.
\squishend 
The output of \algo is a mapping from the
  input feature space to the new representation space that can be
 applied to data records of novel unseen individuals.
\subsection{Contribution}
The key contributions of this paper are:
\squishlist
\item A
practically viable operationalization of the
  individual fairness paradigm that overcomes the challenge of human
  specification of a distance metric, by eliciting easier and more intuitive forms of human judgments.
\item Novel methods for transforming such human judgments into 
pairwise %
constraints in a fairness graph $G$.
\item A mathematical optimization model and representation learning method, called
{\em PFR},
that combines the input data $X$ and the fairness
  graph $G$ into a unified representation by learning a latent model
  with graph embedding.
\item Demonstrating the effectiveness of our approach at achieving
  both individual and group fairness using comprehensive experiments
  with synthetic as well as real-life data on recidivism prediction
  (Compas) and violent neighborhoods prediction (Crime and
  Communities).  
\squishend
\section{Related Work}

\spara{\bf Operationalizing Fairness Notions:} 
Prior works 
on algorithmic
fairness explore two broad families of fairness notions: group
fairness and individual fairness. 

\spara{\bf Group Fairness:}
Two popular notions of group fairness are
demographic parity, which requires equality of beneficial outcome prediction rates between
different socially salient groups, 
\cite{calders2009building,kamiran2010discrimination,pedreshi2008discrimination},
and equalized odds that aims to achieve equality of prediction error rates
between groups
\cite{hardt2016equality}. 
Approaches to achieve 
group fairness include de-biasing the input data via data perturbation,
re-sampling, modifying the value of protected attribute/class
labels
\cite{Salimi2019SIGMOD,kamiran2010discrimination,pedreshi2008discrimination,feldman2015certifying}
as well as incorporating group fairness as an additional
constraint in the objective function of 
machine learning
models 
\cite{kamishima2011fairness,calders2009building,zafar2017AISTATS}.
Similar approaches to achieve group fairness have been proposed for 
fair ranking
 \cite{Asudeh2019SIGMOD,Elbassuoni2O19EDBT,zehlike2017fa}, fair set selection and clustering    
 \cite{chierichetti2017NIPS,Stoyanovich2018EDBT}
Recently, several researchers have highlighted the inherent
incompatibility between different notions of group fairness and the
inherent trade-offs when attempting to achieve them simultaneously
 \cite{kleinberg_et_al:LIPIcs:2017:8156,chouldechova2017fair,friedler2016possibility,corbett2017algorithmic}.

\spara{\bf Bridging Individual and Group Fairness:} 
Approaches to enforcing group fairness have mostly ignored individual
fairness and vice versa.
In \cite{zemel2013learning} and \cite{lahoti2018ifair}, authors operationalize individual fairness by learning a restricted form of distance metric from the data.
Some recent works use the objective of the learning algorithm itself
to implicitly define the similarity metric
 \cite{speicher2018unified,
  Biega:SIGIR2018,kearns2017meritocratic}.  For instance, when learning a
classifier, these works would use the class labels in the training
data or predicted class labels to measure similarity.
However, 
fairness notions are meant to 
address
societal inequities that are not captured in the training data (with potentially
biased labels and missing features).
In such scenarios, the fairness objectives 
are in conflict with 
the learning
objectives.

In this work, we assume that human experts with background knowledge
of past societal unfairness and future societal goals could provide
coarse-grained judgments on whether pairs of individuals deserve
similar outcomes. {\color{black} Other works like \cite{gillen2018online} \cite{jung2019eliciting} make similar arguments.
Further, 
we show that by appropriately constraining
outcomes for pairs of individuals belonging to different groups, we
are able to achieve both individual and group fairness to a large degree. }
\spara{\bf Learning Pairwise Fair Representations:} In terms of our
technical machinery, 
the closest prior work 
is \cite{zemel2013learning,lahoti2018ifair}
that aim to learn new representations for individuals that
``retain as much information in the input feature space as possible,
while losing any information that can identify individuals' protected
group membership''. 
Our approach aims to learn new representations
for individuals that retain the input data to
the best possible extent, while mapping equally deserving individuals as
closely as possible. 
Like \cite{zemel2013learning,lahoti2018ifair} our method can
be used to find representations for new individuals not seen in the
training data. 

Finally, the core optimization problem we formulate 
relates to graph
embedding and representation learning
\cite{hamilton2017representation}. The aim of graph embedding
approaches is to a learn a representation for the nodes in the graph
encoding the edges between nodes as well as the attributes of the
nodes
\cite{lin2005semantic,amid2015multiview}. Similarly, we wish to
learn a representation encoding both the features of individuals as
well as their interconnecting edges in the fairness graph.

 %
%
%
\section{Model}
\label{sec-model}
\subsection{Notation} \label{sec:notation}
\begin{itemize}
\item $X$ is an input data matrix 
{ of $N$ data records and $M$ numerical or categorical attributes.}
We use $X$ to denote both the matrix and the population
of individuals $x_i$:
\begin{equation}
X = [x_1, x_2, x_3, \cdots x_N] \in R^{M \times N} \nonumber
\end{equation} 

\item $Z$ is a low-rank representation of $X$ in 
{\textbf{} a $D$-dimensional space where $D \ll M$.}
\begin{equation}
Z = [z_1, z_2, z_3, \cdots z_N] \in R^{D \times N} \nonumber
\end{equation}

\item $S$ is a 
random variable representing the values that the 
protected-group attribute can take. 
We assume a single attribute in this role; if there are
multiple attributes which require fair-share protection,
we simply combine them into one.
We allow more than two values for this attribute, 
going beyond the usual binary model (e.g., 
gender = male or female, race = white or others). 
$X_s \subset X$ denotes the subset of individuals in $X$ who are members of
group $s \in S$.	

\item $W^X$ is the adjacency matrix of a 
k-nearest-neighbor graph over the input space $X$ given by:
\begin{equation}
W_{ij}^X = 
\begin{cases}
\exp \left(\frac{-\|\mathbf{x_i} -\mathbf{x_j} \|^{2}}{t}\right), \text{if } \mathbf{x_i} \in N_p(x_j) \text{ or } \mathbf{x_j} \in N_p(x_i)\\ \nonumber
0 \hspace{22mm}, \text{otherwise}
\end{cases}
\end{equation}
where $N_p(x_i)$ denotes the set of $p$ nearest neighbors of $x_i$ in Euclidean space (excluding the protected attributes), and $t$ is a scalar hyper-parameter.

\item $W^F$ is the adjacency matrix of the fairness graph $G$ whose nodes  are individuals and whose edges are connections between individuals that are equally
deserving and must be treated similarly.
\end{itemize}

\subsection{From Distance Metric to Fairness Graph}
\label{sec:fairness_graphs}
In this section we address the question of how to elicit 
side-information on individual fairness and model it as a fairness graph $G$ and its corresponding adjacency matrix as $W^F$. The key idea of our approach is rooted in the following observations:
\squishlist
\item Humans have a strong intuition about whether two individuals are similar or not. However, it is difficult for humans to specify a quantitative \emph{similarity metric}.
\item In contrast, it is more natural to make other forms of judgments such as (i)``Is A similar to B with respect to the given task?'', 
or (ii)``How suitable is A for the given task (e.g., on a Likert scale)''.
\item However, these kinds of judgments are difficult to elicit when the pairs of individuals belong to diverse, incomparable groups. In such cases, it is easier for humans to compare individuals within the same group, as opposed to comparing individuals between groups.
Pairwise judgements can be beneficial even if they are available only sparsely,
that is, for samples of pairs.
\squishend

\noindent
Next, we present two  models for constructing fairness graphs, which overcome the outlined difficulties via 
\squishlist
\item[(i)] eliciting (binary) pairwise judgments of individuals who should be treated similarly, or grouping individuals into equivalence classes
(see Subsection \ref{subsec:FGfromEquivalenceClasses})
and 
\item[(ii)] eliciting within-group rankings of individuals and connecting individuals across groups who fall
within the same quantiles of the per-group distributions
(see Subsection \ref{subsec:FGfromGroupQuantiles}).
\squishend

\subsubsection{Fairness Graph for Comparable Individuals}
\label{sec:aggregated-judgments-graph} 
\label{subsec:FGfromEquivalenceClasses}

The most direct way to create a fairness graph is to elicit (binary) pairwise similarity judgments about a small sample of individuals in the %
input data, 
and to create a graph $W^F$ such that there
is an edge between two individuals if they are deemed 
similarly qualified for a certain task (e.g., being
invited for job interviews).

Another alternative is to elicit judgments that map individuals into discrete equivalence classes. 
Given a number of such judgments for a sample of individuals in the input dataset, we can construct a fairness graph $W_F$ by creating an edge between two individuals if they belong to the same equivalence class irrespective of their group membership.

\begin{defn-eqn}{Equivalence Class Graph}{Let $[x_i]$ denote the equivalence class of an element $x_i \in X$. We construct an undirected graph $W^F$ associated to $X$, where the nodes of the graph are the elements of $X$, and two nodes $x_i$ and $x_j$ are connected if and only if $[x_i] = [x_j]$.}
\label{Def:equivalence-class-graph}
\end{defn-eqn}

The fairness graph built from such equivalence classes
identifies equally deserving individuals -- a valuable asset
for learning a fair data representation.
Note that the graph may be sparse, if information on equivalence
can be obtained merely for sampled representatives.
\subsubsection{Fairness Graph for Incomparable Individuals}
\label{sec-between-group-quantile-graph}
\label{subsec:FGfromGroupQuantiles}

However, at times, our individuals are from diverse and incomparable groups. In such cases, it is
difficult if not infeasible to ask humans for pairwise judgments about individuals {\em across groups}.
Even with the best intentions of being fair, human
evaluators may be misguided by wide-spread bias. 
If we can elicit a ranked ordering of individuals per-group, and pool them into quantiles
(e.g., the top-10-percent), then one could  assume that individuals from different groups who belong to the same quantile in their respective rankings,
are similar to each other. 
Arguments along these lines have been made also 
by \cite{kearns2017meritocratic} in their notion of meritocratic fairness. 

Specifically, our idea is to first obtain
within-group rankings of individuals (e.g., rank men and women
separately) based on their suitability for the decision task at hand,
and then construct a between-group fairness graph by linking all
individuals ranked in the same $k^{th}$ quantile across the different
groups (e.g., link programming language researcher and machine learning researcher who are similarly ranked in their own
groups). 
The relative rankings of individuals within a group, whether
they are obtained from human judgments or from secondary data sources, are
less prone to be influenced by discriminatory (group-based)
biases.

Formally, given $(X_s,Y_s)$ for all $s \in S$, where $Y_s$ is a %
random variable depicting the ranked position of 
individuals in $X_s$. We construct a 
{\em between-group quantile graph} 
using Definitions \ref{def-k-quantile} and \ref{def-between-group-quantile-graph}.

\begin{defn-eqn}{k-th quantile}{Given a random variable $Y$, the k-th quantile $Q_k$ is that value of $y$  in the range of $Y$, denoted $y_k$, for which the probability of having a value less than or equal to $y$ is $k$.}
		\begin{equation}
		Q(k)= \{y:Pr(Y \le y)=k\} \quad \text{where} \quad 0<k<1
		\end{equation}
		\label{def-k-quantile}
For the non-continuous behavior of discrete variables,
we would add
appropriate ceil functions to the definition,
but we skip this technicality.
\label{Def:kth-quantile}
\end{defn-eqn}

\begin{defn-eqn}{Between-group quantile graph}{Let
 $X_{s}^{k} \subset X$ denote the subset of individuals who belong to  group $s \in S$ and whose scores lie in the k-th quantile. We can construct a multipartite graph $W^F$ whose edges are given by:
		\begin{equation}
		W_{ij}^F = 
		\begin{cases}
		1 &,\text{ if } x_i \in X_{s}^k \quad  \text{and} \quad x_j \in X_{s'}^k \quad , \quad  s \neq s' \\
		0 &,\text{ otherwise }
		\end{cases}
		\end{equation}
		That is, there exists an edge between a pair of individuals $\{x_i,x_j\} \in X$ if $x_i$ and $x_j$ have different group memberships and their scores $\{y_i,y_j\}$ lie in the same quantile.
For the case of two groups (e.g., gender is male or female),
the graph is a bipartite graph.
	}
	\label{def-between-group-quantile-graph}
\end{defn-eqn}

This model of creating between-group quantile graphs is general enough to consider any kind of per-group ranked judgment.
Therefore, this model is not necessarily limited to legally protected groups (e.g., gender, race), it can be used for any socially salient groups that are incomparable for the given task (e.g., machine learning  vs. programming language researchers).
Note again that the pairwise judgements may be sparse, if such information is
obtained only for sampled representatives.
\subsection{Learning Pairwise Fair Representations}

In this section we address the question: How to encode the background information
such that downstream tasks can make use of it for the decision making?
\subsubsection{Objective Function}

In fair machine learning, such as fair classification models, the 
objective usually is 
to maximize the classifier accuracy 
(or some other quality metric) while satisfying
constraints on group fairness statistics 
such as parity.
For learning fair data representations 
that can be used in any downstream application -- classifiers or regression models with varying target variables unknown at learning time --
the objective needs to be generalized
accordingly.
To this end, the \algo model aims to combine 
the utility of the learned representation and,
at the same time, preserve the information from
the pairwise fairness graph.
{ Starting with 
matrix $X$
of $N$ data records $x_1 \dots x_N$
and $M$ numeric or categorial attributes,
\algo computes a 
lower-dimensional latent matrix $Z$ of $N$ records each with $D < M$ values.}

We
model utility into the notion of preserving 
local neighborhoods of user records in the attribute space $X$
in the latent representation $Z$

Reflecting the fairness graph in the
learner's optimization for $Z$
is a demanding and a priori open problem.
Our solution \algo casts this issue into a
graph embedding that is incorporated into
the overall objective function.
The following discusses the 
technical details of \algo{}'s
optimization.

\spara{Preserving the input data:}
For each data record $x_i$ in the input space,
we consider the set $N_p(x_i)$ of its $p$ nearest neighbors with regard to the distance defined by
the kernel function given by $W_{ij}^X$.
For all points $x_j$ within $N_p(x_i)$, 
we want the corresponding latent representations
$z_j$ to be close to the representation $z_i$,
in terms of their L2-norm distance.
This is formalized by the {\em Loss in $W^X$}, denoted by $Loss_X$.
\begin{equation}\label{eq-utilityPreserving}
Loss_X = \sum\limits_{i,j = 1}^{N} \normF{z_i - z_j}^2 W_{ij}^X
\vspace{-1mm}
\end{equation}
Note that 
this objective
requires only local neighborhoods in $X$ to be preserved in the transformed space. 
We disregard data points outside of 
p-neighborhoods.
This relaxation 
 increases the feasible solution space for the dimensionality reduction.

\spara{Learning a fair graph embedding:} 
Given a fairness graph $W^F$, 
the goal for 
$Z$ is to preserve neighborhood properties in $W^F$. 
In contrast to $Loss_X$, however, we
do not need any distance metric here, but 
can directly leverage the fairness graph.
If two data points $x_i, x_j$
are connected in $W^F$, we 
aim to map them to representations $z_i$
and $z_j$ close to each other.
This is formalized by the {\em Loss in $W^F$}, denoted by $Loss_F$.
\begin{equation}\label{eq-Fairness}
Loss_F = \sum\limits_{i,j = 1}^{N} \normF{z_i - z_j}^2 W_{ij}^F
\end{equation}
Intuitively, for data points connected in $W^F$,
we add a penalty when their representations
are far apart in $Z$.

\spara{Combined objective:}
Based on the above considerations, a fair 
representation $Z$ is computed
 by minimizing the combined objectives of Equations \ref{eq-utilityPreserving} and \ref{eq-Fairness}. 
 The parameter $\gamma$ weighs the importance tradeoff between $W^X$ and $W^F$. 
 As $\gamma$ increases influence of the fairness graph $W^F$ increases.
 An additional orthonormality constraint on $Z$ is imposed to avoid trivial results. The trivial result being that all the datapoints are mapped to same point.
\begin{align} 
\label{eq:combined_objective}
\nonumber
&\text{Minimize} \quad (1 - \gamma)\sum\limits_{i,j = 1}^{N} \|{z_i - z_j}\|^2 W_{ij}^X + \gamma\sum\limits_{i,j = 1}^{N} \|{z_i - z_j}\|^2 W_{ij}^F \\
& \text{subject to} \quad Z^TZ = I
\vspace{-1mm}
\end{align}
\subsubsection{Equivalence to Trace Optimization Problem}
\label{eq-trace-proof}
\noindent
Next, we show that the 
optimization problem in Equation \ref{eq:combined_objective} can be transformed and solved as an equivalent eigenvector problem. 
To do so, we assume that the learnt representation $Z$ is a linear transformation of $X$ given by $Z = V^T X$.

We start by showing that minimizing $\|{z_i - z_j}\|^2 W_{ij}$ is equivalent to minimizing the trace $Tr(V^TXLX^TV)$.
Here we use $W$ to denote $W^X$
or $W^F$, as the following mathematical
derivation holds for both of them analogously:
\begin{align*}
Loss & = \sum\limits_{i,j = 1}^{N} \normF{z_i - z_j}^2 W_{ij} \\ \nonumber
& =   \sum\limits_{i,j = 1}^{N} Tr((z_i - z_j)^T ( z_i - z_j)) W_{ij}  \quad 
\\ \nonumber
& =  2 \cdot Tr(\sum\limits_{i,j = 1}^{N} z_i^T z_i D_{ii} - \sum\limits_{i,j = 1}^{N} z_{i}^Tz_{j} W_{ij}) \\\nonumber
& =  2 \cdot Tr(V^T X L X^T V) \quad 
\nonumber
\end{align*}

\noindent
where $Tr(.)$ denotes the trace of a matrix, $D$ is a diagonal matrix whose entries are column sums of $W$, 
and $L = D-W$ is the graph Laplacian constructed from matrix $W$. 
Analogous to $L$, we use $L^X$ to denote graph laplacian of $W^X$,
and $L^F$ to denote graph laplacian of $W^F$.

\subsubsection{Optimization Problem}
Considering the results of Subsection \ref{eq-trace-proof}, we can transform the above combined objective in Equation \ref{eq:combined_objective} into a trace optimization problem as follows:

\begin{align}
\label{eq-trace-optimization-objective}
\nonumber
&\text{Minimize } J(V) = \quad  Tr\{V^T X ((1 - \gamma)L^X + \gamma L^F) X^T V\} \\
&\text{subject to}  \quad  V^TV = I
\end{align}

\noindent
We aim to learn an 
$M \times D$ 
matrix $V$ such that for each input vector $x_i \in X$, we have the low-dimensional representation $z_i = V^T x_i$, where $z_i \in Z$ is the mapping of the data point $x_i$ on to the learned basis $V$. The objective function is subjected to the constraint $V^TV = I$ to eliminate trivial solutions.

\noindent
Applying Lagrangian multipliers, we can formulate the trace optimization problem in Equation \ref{eq-trace-optimization-objective} as an eigenvector problem
\begin{equation}
\label{eq:original-eigenvector-problem}
X ((1 - \gamma)L^X + \gamma L^F) X^T \mathbf{v_i} = \lambda \mathbf{v_i}
\end{equation}
It follows that the columns of optimal $V$ are the eigenvectors corresponding to $D$ smallest eigenvalues denoted by $V = [\mathbf{v_1}  \mathbf{v_2}  \mathbf{v_3} \cdots \mathbf{v_D}]$, and $\gamma$ is a regularization hyper-parameter. 
Finally, the d-dimensional representation of input $X$ is given by $Z = V^T X$.

\spara{Implementation:} The above standard eigenvalue problem for symmetric matrices can be solved in $O(N^3)$ using iterative algorithms. In our implementation we use the standard eigenvalue solver implementation from scipy.linalg.lapack \\ python library \cite{anderson1990lapack}.

\subsubsection{Inference} { Given an input vector $x_i$ for
a previously unseen individual, the \emph{PFR} method computes its fair representation as $z_i = V^T x_i$ where $z_i$ is the projection of the datapoint $x_i$ on the learned basis $V$. It is important to note that the fairness graph $W^F$ is only required during the training phase to learn the basis $V$. Once the $M \times D$ 
matrix $V$ is learned, we do not need any fairness labels for newly seen data.}

\subsubsection{Kernelized Variants of PFR} 
In this paper, we restrict ourselves to assume that the 
representation $Z$ is a linear transformation of $X$ given by $Z = V^T X$.
However, 
\emph{PFR}
can be generalized to a non-linear setting by replacing $X$ with 
a non-linear mapping $\phi(X)$ 
and then performing \emph{PFR} on the outputs of $\phi$
(potentially in a higher-dimensional space). 

For this purpose, assume that $Z = V^T \Phi(X)$ and $V = \sum\limits_{i = 1}^{n} \alpha_{i}\Phi(x_i)$
with a Mercer kernel matrix $K$ where  $K_{i,j} = k(x_i,x_i) = \Phi(x_i)^T\Phi(x_j)$.
We can show that the trace optimization problem in Equation \ref{eq:original-eigenvector-problem} can be generalized to this non-linear kernel setting, and it can be conveniently solved by working with Mercer kernels without having to compute $\Phi(X)$.
We arrive at the following generalized optimization problem.
\begin{equation}
\label{eq-kernel-optimization-problem}
K ((1 - \gamma)L^X + \gamma L^F) K \mathbf{\alpha_i} = \lambda \mathbf{\alpha_i}
\end{equation}

Analogously to the solution of Equation \ref{eq:original-eigenvector-problem}, the solution to the  \emph{kernel PFR} is given by $A = [\mathbf{\alpha_1}  \mathbf{\alpha_2}  \mathbf{\alpha_3}  \cdots \mathbf{\alpha_D}]$ where $\alpha_1 \cdots \alpha_D$ are the $D$ smallest eigenvectors. 
Finally, the learned representation of $X$ is given by $Z = V^T \Phi(X) = A^T K$.

In this paper we present results only for  \emph{linear PFR},
leaving the investigation of \emph{kernel PFR} for future work.
\section{Experiments}
\label{sec:experiments}
This section reports on experiments with synthetic and real-life datasets. 
We compare a variety of fairness-enhancing methods
on a binary classification task as a downstream application.
We address the following key questions in our main results in Subsection \ref{subsec:synthetic_experiments}, \ref{subsec:crime_experiments} and \ref{subsec:compas_experiments}:
\begin{itemize}
	\item[-] [Q1] What do the learned representations look like?
	\item[-] [Q2] What is the effect on individual fairness?
	\item[-] [Q3] What is the influence on the trade-off between fairness and utility? 
	\item[-] [Q4] What is the influence on group fairness?
\end{itemize}

\noindent
{ In addition, to understand the robustness of our model to the main hyper-parameter $\gamma$, as well as the sensitivity of the model to the number of labels in the fairness graph, we report additional results in Subsection \ref{subsec:influence_of_gamma}, and \ref{subsec:sensitivity_to_labelsize}.}

\subsection{Experimental Setup}
\spara{Baselines:} We compare the performance of 
the following methods
\squishlist
\item {\em Original representation:} a naive representation of the input dataset wherein the protected attributes are masked.
\item {\em iFair \cite{lahoti2018ifair}}: an unsupervised representation learning method, which optimizes for two objectives: (i) individual fairness in $W^X$, and (ii) obfuscating protected %
attributes.
\item {\em LFR \cite{zemel2013learning}}: a supervised representation learning method, which optimizes for three objectives: (i) accuracy (ii) individual fairness in $W^X$ and (iii) demographic parity.
\item {\em Hardt \cite{hardt2016equality}}: a post-processing method that aims 
to minimize the difference in error rates between groups by 
optimizing for the group-fairness measure \emph{EqOdd} (Equality of Odds). 
\item {\em PFR}: Our unsupervised representation learning method
that optimizes for two objectives (i) individual fairness as per $W^F$ and (ii) individual fairness as per $W^X$.
\squishend

\spara{Augmenting Baselines:} For fair comparison we compare \emph{PFR} with augmented versions of all methods (named with \emph{suffix} +). 
In the augmented version, we give each method an advantage by enhancing it with the information in the fairness graph $W^F$. 
Since none of the methods can be naturally extended to 
incorporate the fairness graph as it is, 
we make our best attempt at modeling the fairness labels that are used to construct $W^F$ as additional numerical features in the training data.  { Since we only have judgments for a sample of training data, we treat the rest as missing values and set them to -1. 
Note that this enhancement is only for training data as fairness labels are
not available for unseen test data. This is in line with how PFR uses the pairwise
comparisons: its representation is learned from the training data,
but at test time, only data attributes $X$ are available.
Concrete details for each of the datasets follow in their respective subsections.}

\spara{Hyper-parameter Tuning:} We use the same experimental setup and hyper-parameter tuning techniques 
for all methods.
Each dataset is split into separate training and test sets. 
On the training set, we perform 5-fold cross-validation 
(i.e., splitting into 4 folds for training and 1 for validation) to
find the best hyper-parameters for each model via \emph{grid search}. 
{ Once hyper-parameters are tuned, we use a independent test set to measure performance. All reported results are averages over 10 runs on independent test sets.}

\spara{Datasets and Task:} {We compare all methods on down-stream classification
using three datasets: (i) a synthetic dataset for US university admission with 203 numerical features, and two real-world datasets: (ii) crime and communities dataset for violent neighbourhood predictions with 96 numerical features and 46 one-hot encoded features (for categorical attributes), and (iii) compas dataset for recidivism prediction with 9 numerical and 420 one-hot encoded features. In order to check the \emph{``true''} dimensionality of the datasets we computed the 
smallest rank $k$ for SVD that achieves a relative error of at most 0.01 for the Frobenius norm
difference between the SVD reconstruction and the original data.
For the three datasets, these dimensionalities are
156, 69, and 117 respectively.}
Table \ref{tbl:dataset-statistics} summarizes the statistics for each dataset, including 
base-rate (fraction of samples belonging to the positive class, for both the protected group and
its complement). 
In all experiments, the representation learning methods are followed by an out-of-the-box logistic regression classifier trained on the corresponding representations. 

\begin{table}[tbh!]
	\center \setlength\tabcolsep{2 pt}
	\caption{{Experimental settings and dataset statistics}}
	\label{tbl:dataset-statistics}
	\vspace{-2mm}
	\noindent\resizebox{\linewidth}{!}{%
		\begin{tabular}{lcHHccccHc}
			\toprule
			Dataset & No of. & $\norm{X_{s=0}}$ & $\norm{X_{s=1}}$ & No. of & True & Base-rate &  Base-rate & Classification & Protected \\
			& records & &  & features & Rank & (s = 0) & (s = 1) & task & attribute \\
			\midrule
			Synthetic &        1000 &                  500 &                  500 &             203 &        156    &    0.51 &                0.48 & Is successful & Race\\
			Crime &       1993 &                 1423 &                  570 &           142 &         69 &       0.35 &                0.86  & Is violent & Race\\
			Compas &       8803 &                 4218 &                 4585 &            429  & 117&                0.41 &                0.55  & Is rearrested & Race\\
			\bottomrule
		\end{tabular}}
	\end{table}
	
\spara{Evaluation Measures:} 
\label{section:fairness-measures}
\squishlist
\item \textbf{Utility} is measured as AUC (area under the ROC curve).
\item \textbf{Individual Fairness} is measured as the \emph{consistency} of outcomes between individuals who are similar to each other. We report consistency values as per both the similarity graphs, $W^X$ and $W^F$.
\begin{equation*}
Consistency = 1 - \frac{\sum \limits_i \sum \limits_j \norm{\hat{y}_i - \hat{y}_j} \cdot W_{ij}}{\sum \limits_i \sum \limits_j W_{ij}} \quad \forall \quad i\neq j
\end{equation*}
\item \textbf{Group Fairness} 
\squishlist
\item \textbf{Disparate Mistreatment (aka. Equality of Odds):} A binary classifier avoids
disparate mistreatment if the group-wise error rates are the same across all groups.
In our experiments, we report per-group false positive rate (FPR) and false negative rate (FNR).
\item \textbf{Disparate Impact (aka. Demographic Parity):} A binary classifier avoids disparate impact if the rate of positive predictions is the same across all groups $s \in S$:
\begin{equation}
P (\hat{Y} = 1 | s = 0) = P (\hat{Y} = 1 | s = 1)
\end{equation}
In our experiments, we report per-group rate of positive predictions.

\squishend
\squishend

\subsection{Experiments on Synthetic Data}
\label{subsec:synthetic_experiments}

\begin{figure*}[tbh!]
	\begin{subfigure}{0.245\linewidth}
		\centering	
		\includegraphics[scale=0.46]{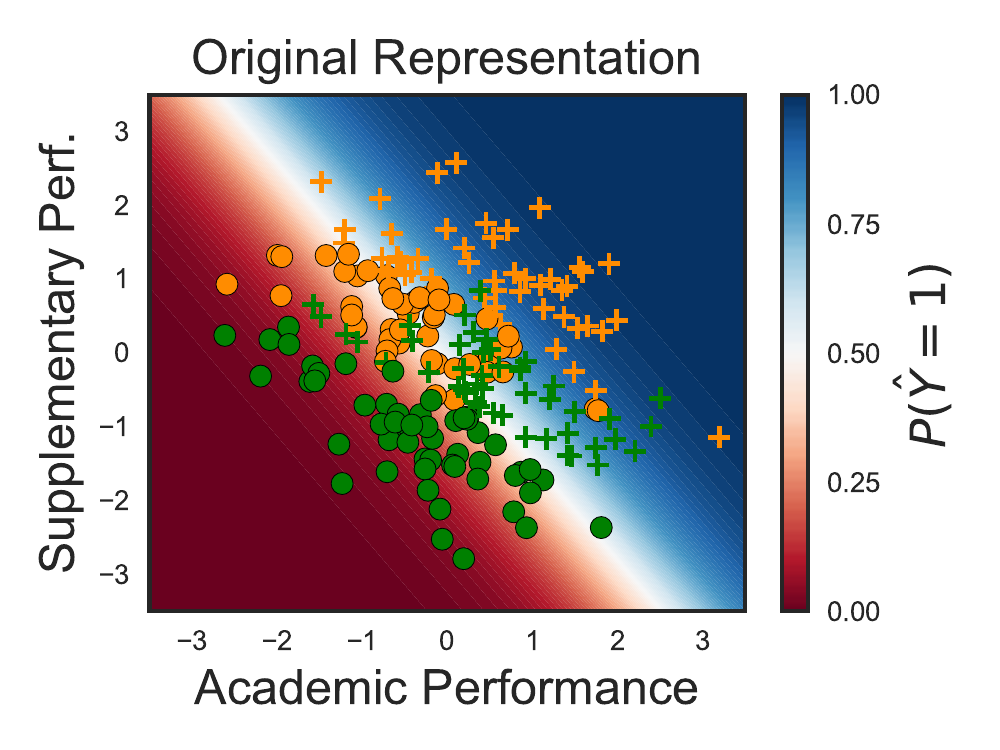}
		\caption{\emph{Original+} Representation}
		\label{fig:synthetic_data_original}
	\end{subfigure}
	\begin{subfigure}{0.245\linewidth}
		\centering	
		\includegraphics[scale=0.46]{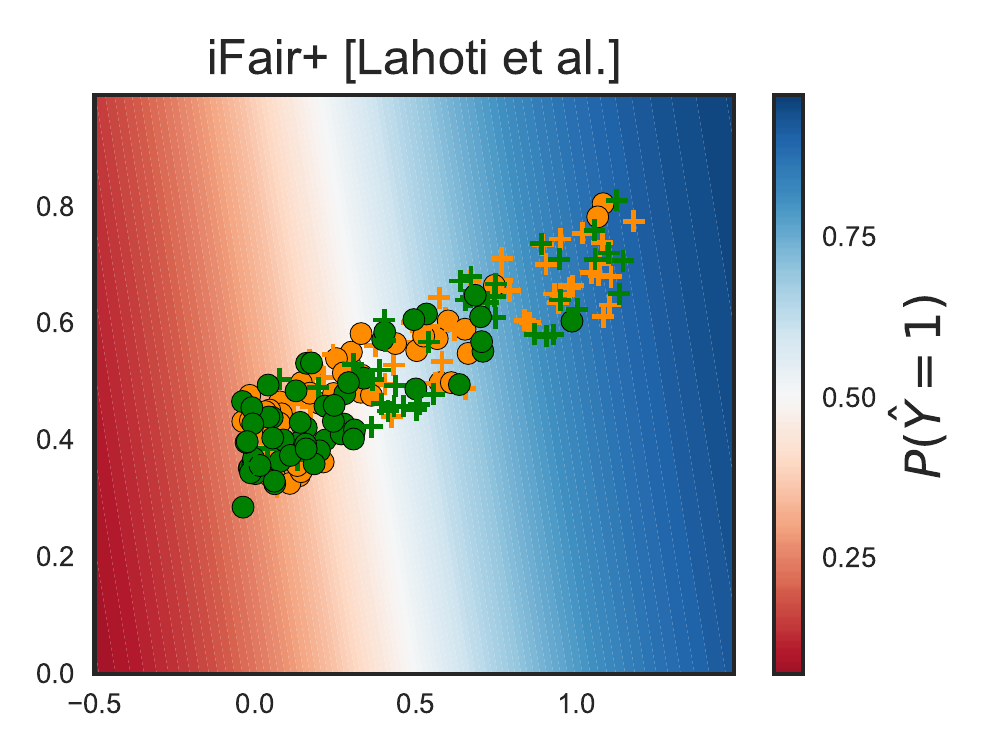}
		\caption{\emph{iFair}+ Representation}		
		\label{fig:synthetic_data_iFair}
	\end{subfigure}			
	\begin{subfigure}{0.245\linewidth}
		\centering	
		\includegraphics[scale=0.46]{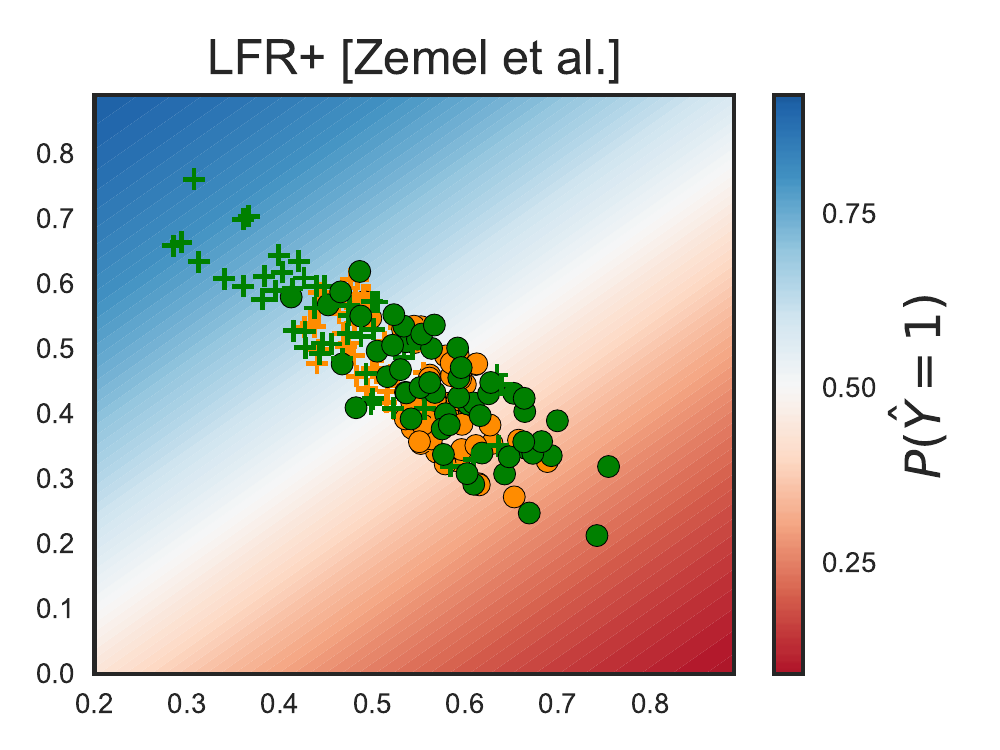}
		\caption{\emph{LFR}+ Representation}		
		\label{fig:synthetic_data_LFR}
	\end{subfigure}	
	\begin{subfigure}{0.20\linewidth}
		\centering	
		\includegraphics[scale=0.45]{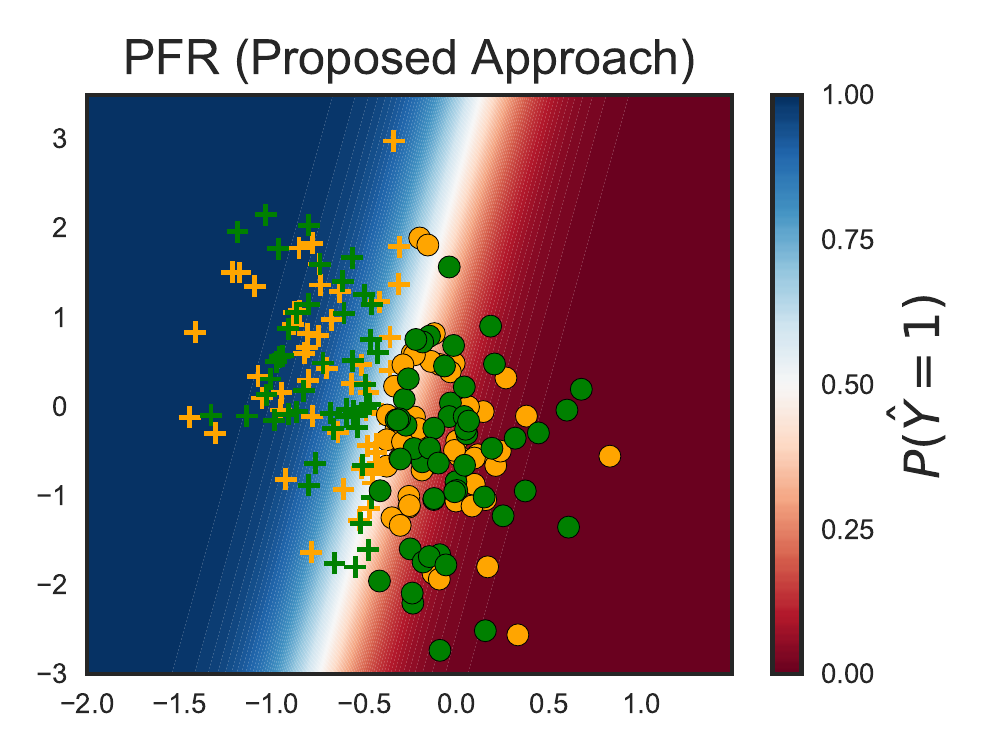}
		\caption{PFR Representation}		
		\label{fig:synthetic_data_PFR}
	\end{subfigure}	
	\caption{{Comparison of (a)\emph{Original+} representation (b)\emph{iFair+} (c)\emph{LFR+} and (d)\emph{PFR} representations on a synthetic dataset. 
		Colors depict membership to protected group (S): orange (non-protected) and green (protected). Markers denote \emph{true} class labels: $Y=1$ (marker +) and $Y=0$ (marker o). Contour plots visualize decision boundary of a classifier trained on the representations. Blue color corresponds to predicted positive classification, red to predicted negative class.
	}}
	\label{fig:learned_representations_Synthetic}
	\vspace{-2mm}
\end{figure*}

\begin{figure}[tbh!]
	\centering	
	\includegraphics[scale=0.52]{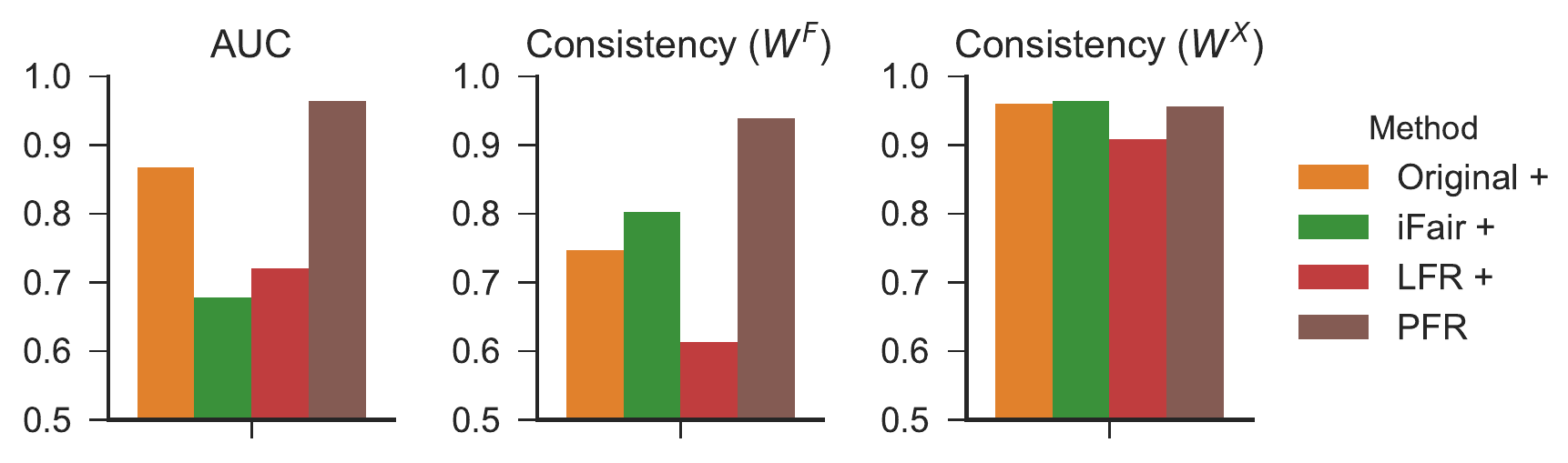}
	\vspace{-3mm}%
	\caption{Results for Synthetic low dimension dataset: Comparison of utility vs individual fairness trade-off across methods. Higher values are better.}	
	\label{fig:Optim-consistency_AUC_WX_WF_Synthetic}
	\vspace{-4mm}
\end{figure}

{
We simulate the US graduate admissions scenario of Section \ref{sec-motivation} where our task is it to predict the ability of a candidate to complete graduate school (binary classification). 
To this end, we imagine that the features in a college admission task can be grouped into two categories. First set of features which are related to their \emph{academic performance} such as overall GPA, grades in each of the high schools subjects like  Mathematics, Science, Languages, etc. 
Second set of features are related to their \emph{supplementary performance} which constitute their overall application package such as SAT scores, admission essay, extracurricular activities, etc. 

We assume that the scores for the second set of features can be inflated for individuals who have higher privilege in the society, for instance by re-taking SAT exam, and receiving professional coaching. 
Suppose we live in a society where our population consists of two groups $s =$ 0 or 1, and the group membership has a high correlation with individual's privilege. This would result in a scenario where the two groups have different feature distributions. Further, if we assume that the inflation in the scores does not increase the ability of the candidate to complete college, the relevance functions for the two groups would also be different.
}

\spara{Creating Synthetic Datasets:} 
{%
We simulate this scenario by generating synthetic data for two population groups $X_0$ and $X_1$. Our dataset consists of three main features: group, \emph{academic performance}, and \emph{supplementary performance}. 
The correlation between \emph{academic performance} and \emph{supplementary performance} is set to 0.3. 
We have additional 100 numerical features with high correlation to \emph{academic performance}, and 100 numerical features with high correlation to \emph{supplementary performance}.
We set the value of correlation between related features by drawing uniformly from [0.75, 1.0].
We use the correlation between features to construct the covariance matrix for a multivariate Gaussian distribution of dimensionality 203.
To reflect the point
that one groups has inflated scores for the features related to \emph{supplementary performance}, we set the mean for these features for the non-protected group one standard deviation higher than the mean for the protected group.

In total we generate 600 samples for training, and 400 samples 
as a withheld test set. We run our experiments on two versions of the synthetic dataset: (i) a
{\em low-dimensional} dataset, which is a subset of the high-dimensional data 
consisting of only three features: Group, Academic Performance and Supplementary performance, and (ii) a {\em high-dimensional} dataset with all 203 features. Experiments on the low-dimensional dataset are performed in order to be able to visually compare the original and learned representations. 
Dataset statistics are shown in Table \ref{tbl:dataset-statistics}.
}

\spara{Ground Truth Labels:} 
{Despite average score on \emph{supplementary performance} features for group $X_{s = 0}$ being higher than for the protected group $X_{s = 1}$, 
we assume that the ability to complete graduate school is the same for both groups;
that is, members of $X_{s = 0}$ and $X_{s = 1}$  are equally deserving
if we adjust their \emph{supplementary performance} scores.
To implement this scenario, 
we set the \emph{true} class label for group $X_{s = 1}$ to positive (1) 
if  \emph{academic} + \emph{supplementary} score $\geq 0$ and for group $X_{s = 0}$ as positive (1) if \emph{academic} + \emph{supplementary} score $\geq 1$. 
Figure \ref{fig:synthetic_data_original} visualizes the generated dataset. 
The colors depict the  membership to groups (S): S = 0 (orange) and S = 1 (green).
The markers denote \emph{true} class labels Y = 1 (marker +) and Y = 0 (marker o). 
}

\spara{Fairness Graph $W^F$:} {In this experiment we simulate the scenario for eliciting human input on fairness, wherein we have access to a fairness oracle who can make the judgments of the form ``Is A similar to B?'' described in Subsection \ref{subsec:FGfromEquivalenceClasses}. To this end, we randomly sample $N\,log_2\,N:= 5538$ pairs (out of the possible $N^2:=600 \times 600$).
We then constructed ours fairness graphs $W^F$ by querying a fairness oracle for Yes/No answers to similarity pairs. If the two points are similar we add an edge between the two nodes.

Fairness oracle for this task is a machine learning model consisting of two separate logistic regression models, one for each group, $X_{S = 0}$ and $X_{S = 1}$ respectively. Given a pair of points, if their prediction probabilities fall in the same quantile, they are deemed similar by the fairness oracle.}

\spara{Augmenting Baselines:} 
{%
We 
cast each row of the matrix $W^F$ (of the fairness graph) 
into $n$ additional binary features for the respective individual.
That is, for every user record, $n$ additional 0/1 features
indicate pairwise equivalence. 
All baselines have access to 
this information via the augmented input matrix $X$. 
}

\subsubsection{Results on Synthetic Low Dimension Dataset}
\label{subsec:lowdimsynth}
\spara{[Q1] What do the learned representations look like?} 
In this subsection we inspect the original representations  and contrast them with learned representations via {\emph{iFair+} \cite{lahoti2018ifair}, \emph{LFR+} \cite{zemel2013learning}}, and our proposed model \emph{PFR}.
Figure \ref{fig:learned_representations_Synthetic} visualizes the original dataset and the
learned representations for each of the models with the number of latent dimensions set to $d = 2$ during the learning.
The contour plots in (b), (c) and (d)
denote the decision boundaries of logistic regression classifiers trained on the respective learned representations. 
Blue color corresponds to positive classification, red to negative;
the more intensive the color, the higher or lower the score of the classifier.
We observe several interesting points:
\squishlist
\item 
First, in the original data, the two groups are separated from each other:  \emph{green} and \emph{orange} datapoints are relatively far apart.
Further, the deserving candidates of one group are relatively far away from the deserving candidates of the other group. That is, ``green plus'' are far from ``orange plus'',
illustrating the inherent unfairness in the original data.
\item In contrast, for all three representation learning techniques -- {\emph{iFair+}, \emph{LFR+}} and \emph{PFR} -- the \emph{green} and \emph{orange} data points are well-mixed. 
This shows that these representations are able to make protected and non-protected group members indistinguishable from each other -- a key property towards fairness.
\item The major difference between the learned representations is that \emph{PFR} 
succeeds in mapping the deserving candidates of one group 
close to the deserving candidates of the other group
(i.e., ``green plus'' are close to ``orange plus''). 
Neither {\emph{iFair+} nor \emph{LFR+}} can achieve this desired effect to the same extent.
\squishend

\spara{[Q2] Effect on Individual Fairness:}
Figure \ref{fig:Optim-consistency_AUC_WX_WF_Synthetic} shows the best achievable trade-off between utility and the two notions of individual fairness.
\squishlist
\item Individual fairness regarding $W^F$: 
We observe that \emph{PFR} significantly outperforms all competitors in terms of \emph{consistency} ($W^F$). 
This follows from the observation that, unlike \emph{Original+}, \emph{iFair+} and \emph{LFR+} representations, \emph{PFR} maps similarly deserving individuals close to each other
in its latent space.
{\item Individual fairness regarding $W^X$: We observe that \emph{PFR}'s performance for  \emph{consistency} ($W^X$) is as good as other approaches, however \emph{PFR} manages to achieve high performance for significantly better performance on \emph{AUC} and \emph{consistency} ($W^F$).}

\squishend

\spara{[Q3] Trade-off between Utility and Fairness:}
The \emph{AUC} bars in Figure \ref{fig:Optim-consistency_AUC_WX_WF_Synthetic} 
show the results on classifier utility for the different methods under comparison.
\squishlist
\item Utility (AUC): PFR achieves by far the best \emph{AUC}, even outperforming the original representation.
While this may surprise on first glance, it is indeed an expected outcome. 
{The fairness edges in $W_F$ help \emph{PFR} overcome the challenge of different groups having different feature distributions (observe Figure \ref{fig:synthetic_data_original}). In contrast, \emph{PFR} is able to learn a unified representation that maps deserving candidates of one group close to deserving candidates of the other group (observe Figure \ref{fig:synthetic_data_PFR}), which helps in improving \emph{AUC}.}
\squishend

\spara{[Q4] Influence on Group Fairness:}
In addition to 
\emph{Original+}, \emph{iFair+}, \emph{LFR+} and PFR, 
we include 
the \emph{Hardt} model in the comparison here,
as it is widely viewed as the state-of-the-art method for group fairness.

\begin{figure}[tbh!]	
	\begin{subfigure}{0.99\columnwidth}
		\centering	
		\includegraphics[scale=0.53]{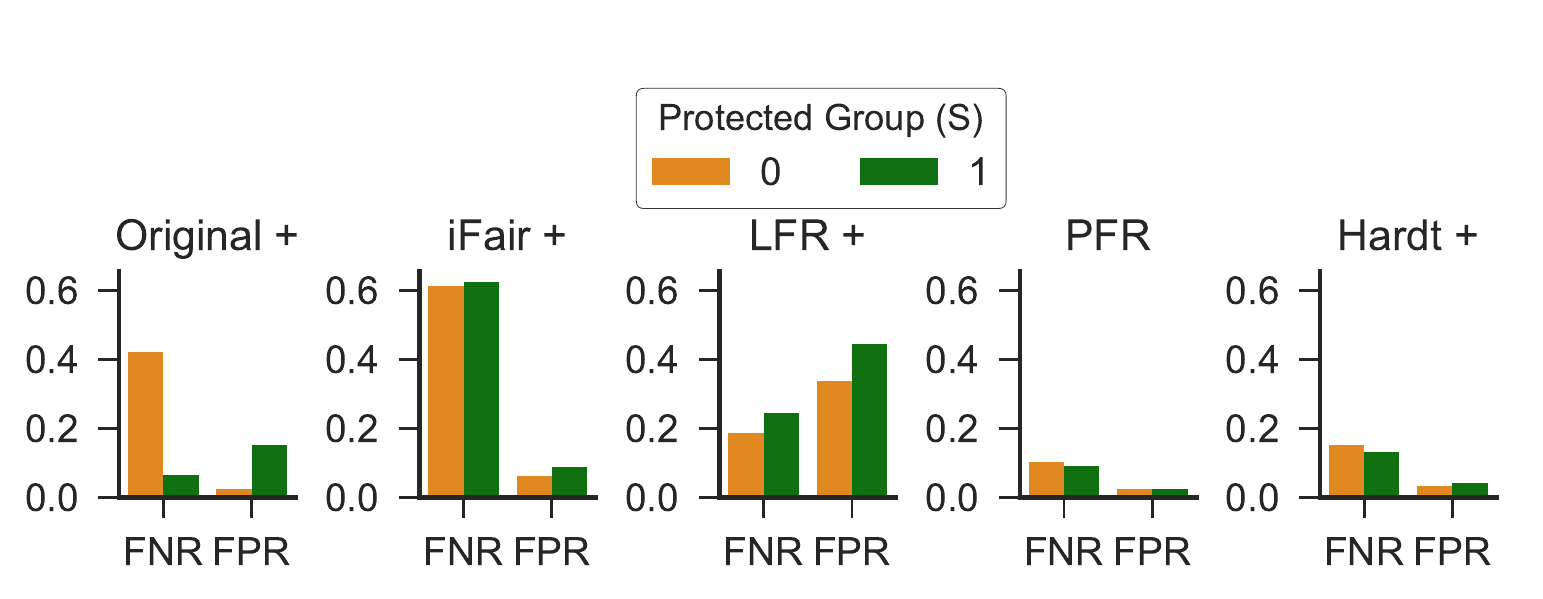}
		\caption{Difference in error rates (FPR and FNR)}
		\label{fig:FPR_FNR_Synthetic}
	\end{subfigure}
		\newline
	\begin{subfigure}{0.99\columnwidth}
		\centering	
		\includegraphics[scale=0.53]{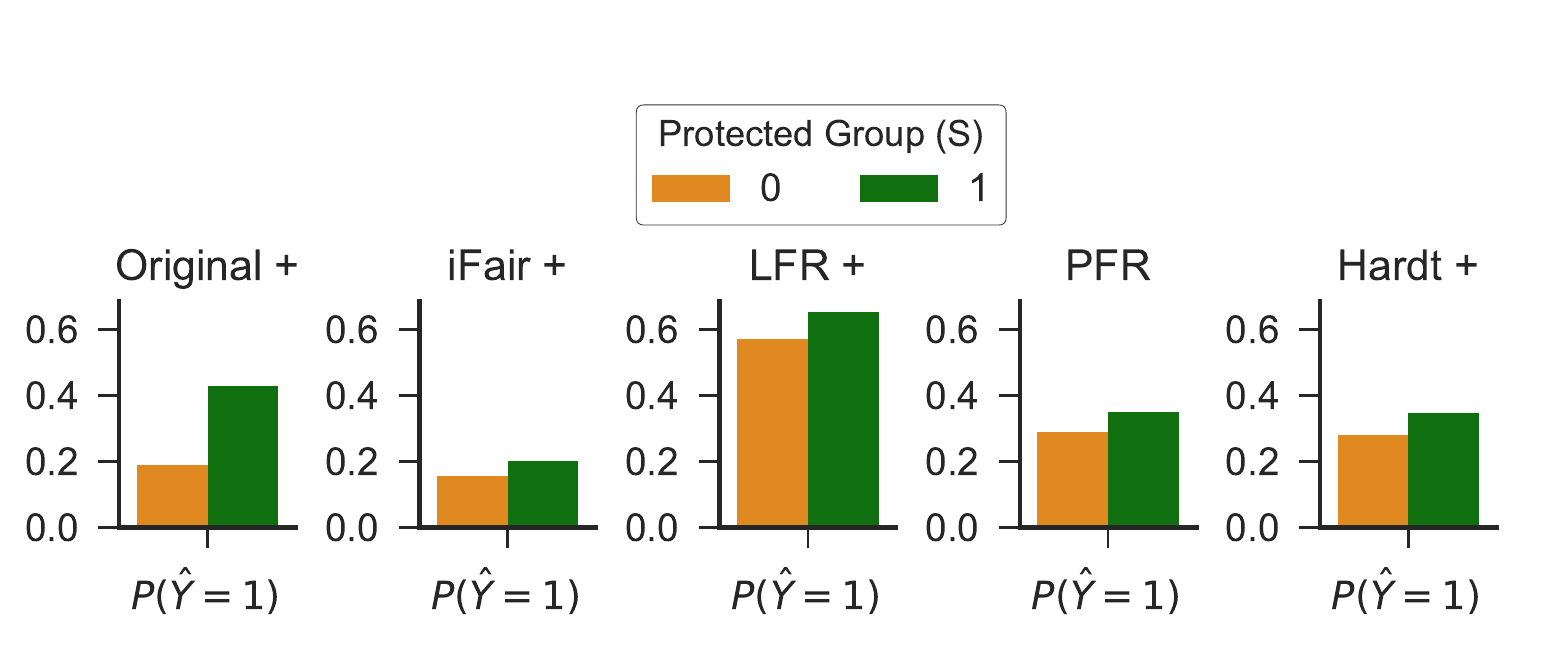}
		\caption{Difference in rates of positive prediction}
		\label{fig:Parity_Synthetic}
	\end{subfigure}
	\caption{Results for Synthetic low dimension dataset: Difference in (a) error rates and (b) rate of positive predictions between protected and non-protected groups  }
	\label{fig:GroupFairness_Synthetic}
\end{figure}

Figure \ref{fig:FPR_FNR_Synthetic} shows the per-group error rates, and Figure \ref{fig:Parity_Synthetic} shows the per-group positive prediction rates. The smaller the difference in the values of the two groups, the higher the group fairness.
We make the following interesting observations:
\squishlist
\item Disparate Mistreatment (Figure \ref{fig:FPR_FNR_Synthetic}): 
We observe that \emph{Original+} model has high difference in error rates (aka. Equality of Odds). 
{\em iFair+} and {\em LFR+} balance the error rates across groups fairly well,
but still have fairly high error rates, indicating their loss on utility.
{\em PFR} and {\em Hardt} have well balanced error rates and generally
lower error. 
For {\em Hardt}, this is the expected effect, as it is optimized for the very goal
of Equality of Odds.
{\em PFR} achieves the best balance and lowest error rates, which is
remarkable as its objective function does not directly consider group fairness.
Again, the effect is explained by {\em PFR} succeeding in mapping
equally deserving individuals from both groups to close proximity in
its latent space.
\item Disparate Impact (Figure \ref{fig:Parity_Synthetic}): The \emph{Original+} approach exhibits a substantial difference in the per-group positive predictions rates of the two groups.
{In contrast, \emph{iFair+}, \emph{LFR+}, and \emph{PFR} representation have the \emph{orange} and \emph{green} data points well-mixed, and this way achieve nearly equal rates of positive predictions
for both groups. Likewise \emph{Hardt+} has the same desired effect.}
\squishend

\subsubsection{Results on Synthetic High Dimension Dataset}
{The results for the high-dimensional synthetic data are largely consistent with the results for 
the low-dimensional case of Subsection  \ref{subsec:lowdimsynth}. 
Therefore, we discuss them only briefly.
Figure \ref{fig:Optim-consistency_AUC_WX_WF_Synthetic_II}
shows results for \emph{AUC}, \emph{consistency}($W^F$),
and \emph{consistency}($W^X$).
Figure \ref{fig:GroupFairness_Synthetic_II} shows results on group fairness measures.

\spara{Utility vs. Individual fairness} regarding $W^F$: 
On first glance, \emph{LFR+} seems to perform best on 
consistency with regard to $W^F$. However, this is trivially
achieved by giving the same prediction to almost all
datapoints: the classifier using the learned \emph{LFR+}
representation accepts virtually all individuals, hence its very poor \emph{AUC}
of around 0.55.
In essence, \emph{LFR+} fails to learn how to cope with
the utility-fairness trade-off. Therefore, we consider this method
as degenerated (for this dataset) and dismiss it as a real baseline.

Among the other methods, \emph{PFR} significantly outperforms all competitors by achieving the best performance on \emph{consistency} ($W^F$), similar performance  as other approaches on \emph{consistency} $(W^X)$, but for a significantly better performance on \emph{AUC}, as shown in Figure \ref{fig:Optim-consistency_AUC_WX_WF_Synthetic_II}. 

\spara{\bf Group Fairness:} Once again, \emph{PFR} clearly outperforms all other methods on group fairness. It achieves near-equal error rates across groups, and  near-equal rates of positive predictions as shown in Figures \ref{fig:FPR_FNR_Synthetic_II} and \ref{fig:Parity_Synthetic_II}. 
Again, \emph{PFR}'s performance on group fairness is as good as that of \emph{Hardt} which is solely designed for equalizing error rates by post-processing the classifier's outcomes.
\emph{LFR+}  seems to achieve good results as well,
but this is again due to accepting virtually all individuals (see above).

\begin{figure}[tbh!]
	\centering	
	\includegraphics[scale=0.495]{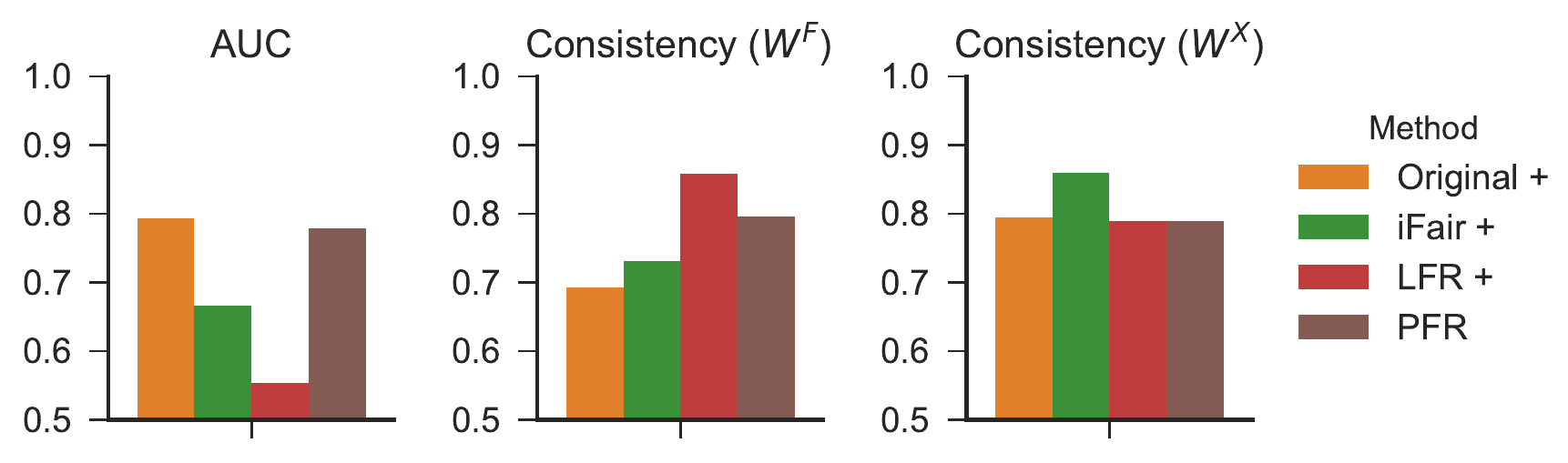}
	\vspace{-3mm}%
	\caption{Results for Synthetic high dimension dataset: Comparison of utility vs individual fairness trade-off across methods. Higher values are better.}	
	\label{fig:Optim-consistency_AUC_WX_WF_Synthetic_II}
	\vspace{-4mm}
\end{figure}

\begin{figure}[h!]	
		\begin{subfigure}{0.99\columnwidth}
			\vspace{-1mm}
			\centering	
			\includegraphics[scale=0.55]{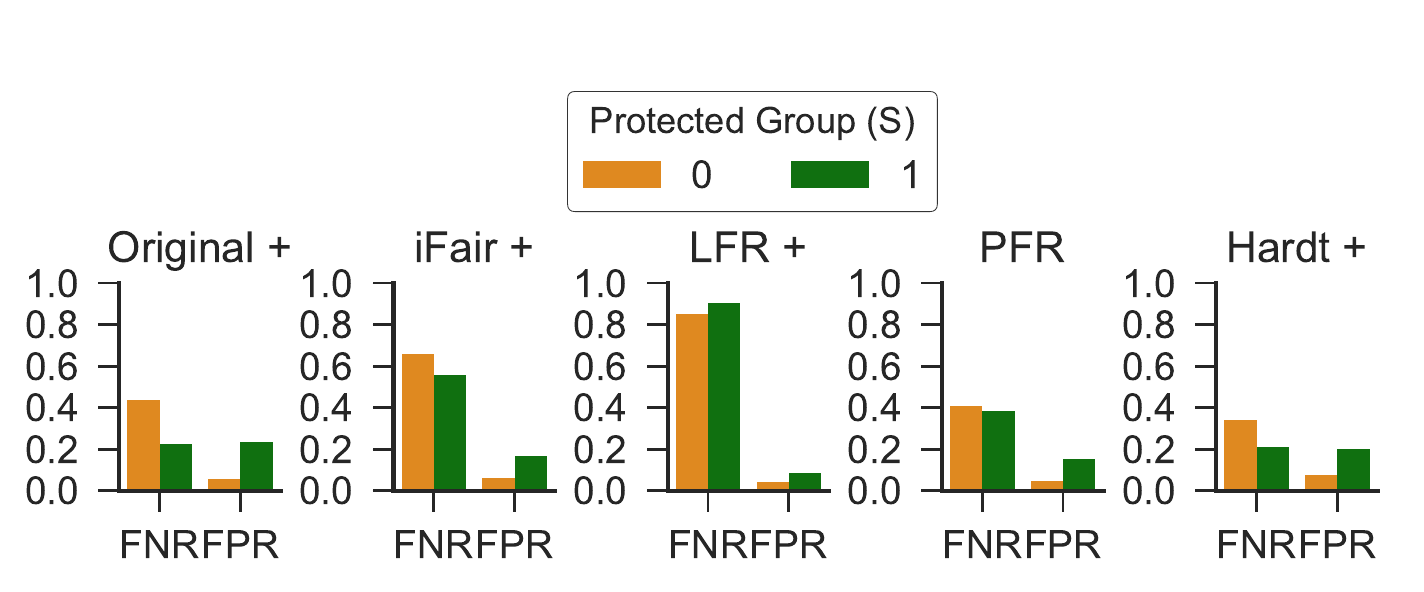}
			\caption{Difference in error rates (FPR and FNR)}
			\label{fig:FPR_FNR_Synthetic_II}
			\end{subfigure}
			\newline
			\begin{subfigure}{0.99\columnwidth}
				\centering	
				\includegraphics[scale=0.55]{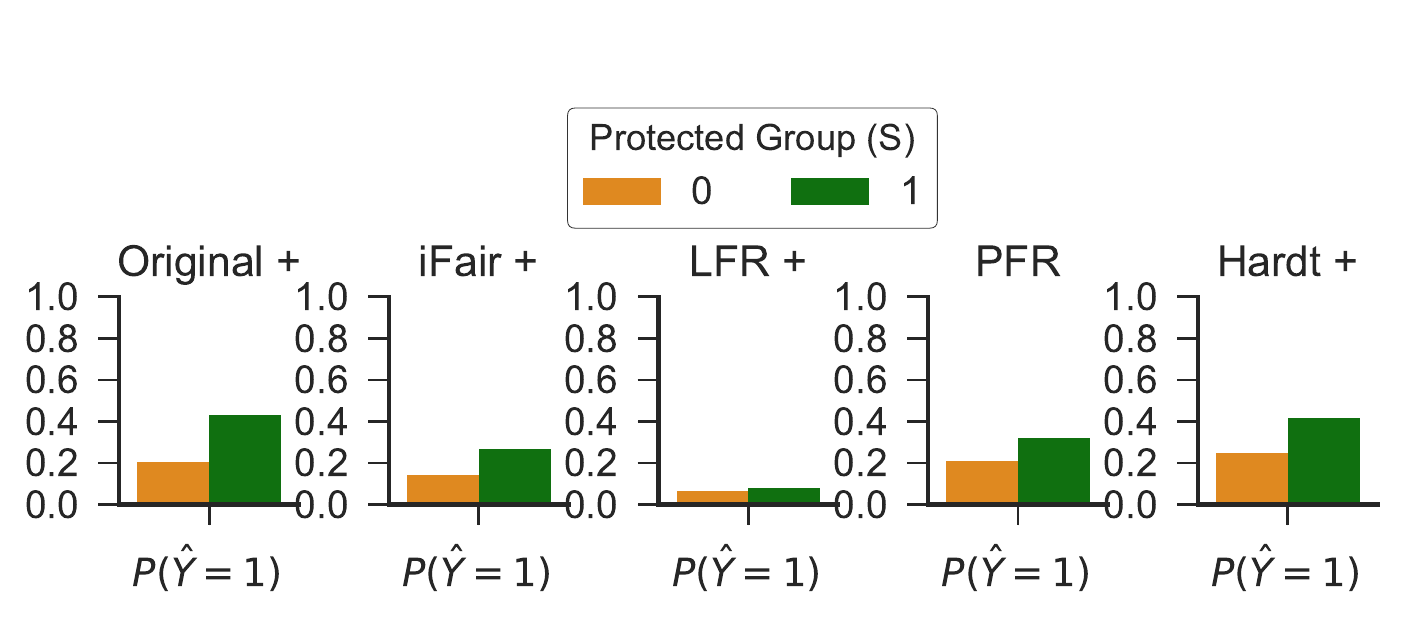}
				\caption{Difference in rates of positive prediction}
				\label{fig:Parity_Synthetic_II}
			\end{subfigure}
\vspace{-0.1cm}
			\caption{Results for Synthetic high dimension dataset: Difference in (a) error rates between protected and non-protected groups and (b) rate of positive predictions.}
			\label{fig:GroupFairness_Synthetic_II}
\end{figure}
}

\subsection{Experiments on Real-World Datasets}
\label{subsec:realworld_experiments}

We evaluate the performance of \algo on the following two real world datasets 
\squishlist
	\item {\em Crime \& Communities}  \cite{crimecommunities2009} is a dataset 
	consisting of socio-economic (e.g., income), demographic (e.g., race), and law/policing data (e.g., patrolling) records for neighborhoods in the US. We set 
	\emph{isViolent} as target variable for  a binary classification task. 
	We consider the communities with majority population white as non-protected group and the rest as protected group.
	\item {\em Compas} data collected by ProPublica \cite{angwin2016machine} contains criminal records comprising offenders' criminal histories and
	demographic features (gender, race, age etc.).
	We use the information on whether the offender was re-arrested as the target variable for binary classification.
	As protected attribute $s \in \{0,1\}$ we use race: African-American (1) vs. others (0).
\squishend

\subsubsection{Constructing the Fairness Graph $W^F$} 
\spara{Crime \& Communities:} We need to elicit pairwise judgments of similarity 
that model whether two neighborhoods are similar in terms of {crime and safety}. 
To this end, we collected human reviews on {crime and safety} for neighborhoods in the US from \url{http://niche.com}. 
The judgments are given in the form of 1-star to 5-star ratings by current and past residents of these neighborhoods. We aggregate the judgments and compute mean ratings for all neighborhoods. 
We were able to collect reviews for about 1500 (out of 2000) communities. 
$W^F$ is then constructed by the technique of Subsection \ref{sec:aggregated-judgments-graph}.
Although this kind of human input is subjective, the aggregation over many reviews
lifts it to a level of inter-subjective side-information reflecting social consensus
by first-hand experience of people.
Nevertheless, the fairness graph may be biased in favor of the African-American neighborhoods,
since residents
tend to have positive perception of their neighborhood's safety.

\spara{Compas:} 
We need to elicit pairwise judgments of similarity 
that model whether two individuals are similar in terms of deserving
to be {granted parole} and not becoming re-arrested later.
However, it is virtually impossible for a human judge 
to {fairly} compare 
people from the groups of
\emph{African-Americans} vs. 
\emph{Others}, without imparting 
the historic bias.
So this is a case, where 
we need to elicit pairwise  judgments between diverse and incomparable groups.

We posit that it is fair, though, to elicit {\em within-group} rankings of risk assessment 
for each of the two groups, to create edges between individuals 
who belong to the same risk quantile of their respective group. 
To this end, we use Northpointe's Compas {decile scores} \cite{brennan2009evaluating} as  background information about within-in group ranking.
These \emph{decile scores} are computed by an undisclosed commercial algorithm which takes as input official criminal history and interview/questionnaire answers to a variety of behavioral, social and economic questions 
(e.g., substance abuse, 
school history, family background etc.).
The decile scores assigned by this algorithm are {\em within-group} {scores} and are not meant to be  compared across groups. 

{We sort these decile scores for each group seprately to simulate per-group ranking fairness judgments.
We then use these per-group rankings as the fairness judgment to construct the fairness graphs for incomparable individuals as discussed in Subsection \ref{subsec:FGfromGroupQuantiles}. Specifically, we compute $k$ quantiles over the ranking as per Definition \ref{def-k-quantile} and then,
construct $W^F$ as described in Definition \ref{def-between-group-quantile-graph}.
}Note that this fairness graph has an implicit anti-subordination assumption. That is, it assumes that individuals in k-th risk quantile of one group are similar to the individuals in k-th quantile of other group - irrespective of their true risk.

{\spara{Augmenting Baselines:} We give our baselines access to the elicited fairness labels by adding them as numerical features to the rows of the input matrix $X$. 
For the  Crime and Communities data, 
we added the elicited ratings (1 to 5 stars) as numerical features, with missing values set to -1. 
For the Compas data, where the fairness labels are per-group rankings, we added the ranking position of each individual within its respective 
group as a numerical feature.}

\subsubsection{Results on Crime \& Communities Dataset}
\label{subsec:crime_experiments}
\begin{figure}[tbh!]
	\centering	
	\includegraphics[scale=0.55]{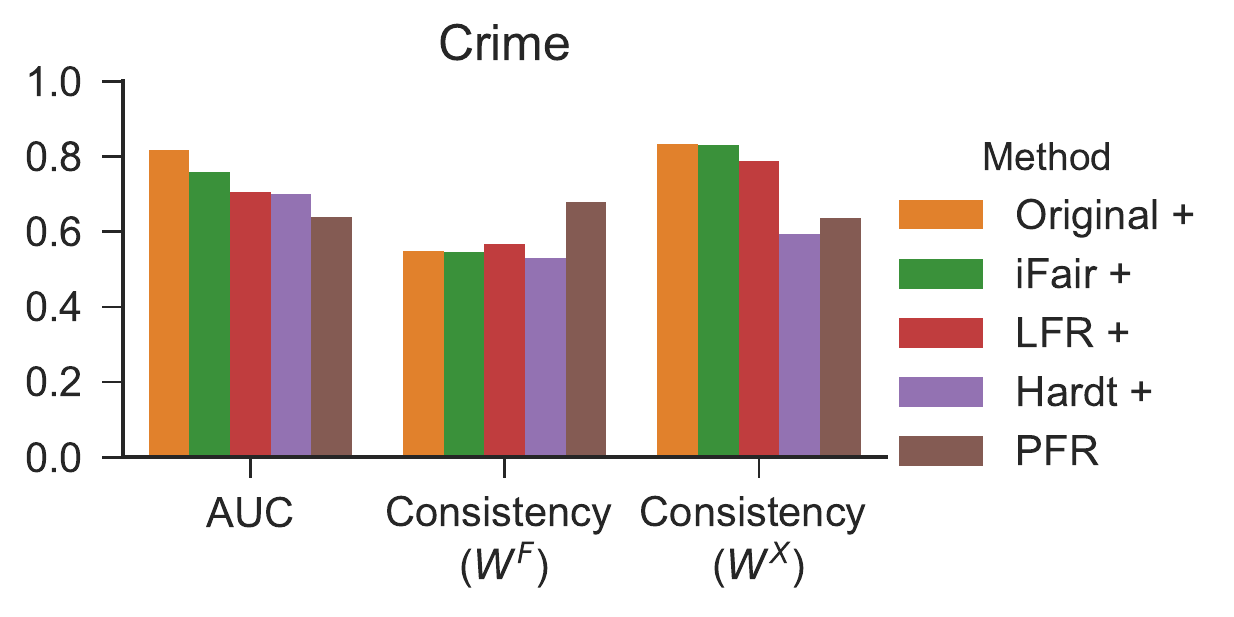}
\vspace{-0.4cm}
	\caption{Crime \& Communities data: utility vs. individual fairness (higher is better).}
	\label{fig:Optim-consistency_AUC_WX_WF_Crime}
	\vspace{-2mm}
\end{figure}

\spara{[Q2] Effect on Individual Fairness:}
Results on individual fairness and utility are given in
Figure \ref{fig:Optim-consistency_AUC_WX_WF_Crime}. 
{We observe that even though all the methods have access to the same fairness information, only \emph{PFR} shows an improvement in consistency $W^F$ over the baseline.} \emph{PFR} outperforms all other
methods on individual fairness (consistency $W^F$). 
However, this gain for  $W^F$ comes at the cost of losing in consistency as per $W^X$. 
So in this case, the pairwise input from human judges exhibits pronounced tension
with the data-attributes input.
Deciding which of these sources should take priority is a matter of
application design.

\spara{[Q3] Trade-off between Utility and Fairness:}
The higher performance of \emph{PFR} on individual fairness regarding
$W^F$ comes with a drop in utility as shown by the AUC bars in Figure \ref{fig:Optim-consistency_AUC_WX_WF_Crime}.
This is because, unlike the case of the synthetic data in Subsection \ref{subsec:synthetic_experiments}, the side-information for the fairness graph $W^F$ 
is not strongly aligned with the ground-truth for the classifier.
{In terms of relative comparison, we observe that only \emph{PFR} shows an improvement in consistency $W^F$ over the baseline, the other approaches show no improvement. The performance of \emph{iFair+} and \emph{LFR+} on consistency on $W^F$ and consistency on $W^X$ is same as that of \emph{Original+}, however for a lower \emph{AUC}. \emph{Hardt+} loses on all the three measures.}

\begin{figure}[tbh!]	
	\begin{subfigure}{0.99\columnwidth}
		\vspace{-1mm}
		\centering	
		\includegraphics[scale=0.55]{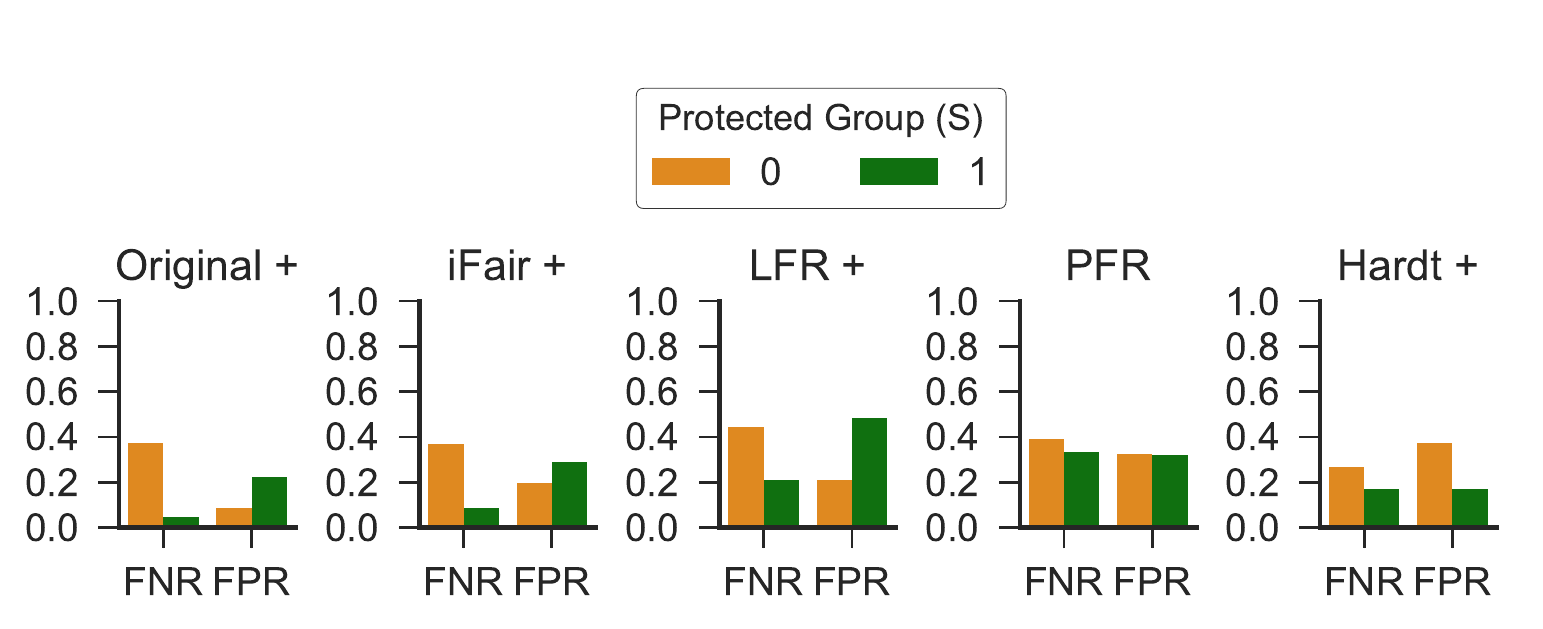}
\vspace{-0.3cm}
		\caption{Difference in per-group error rates (FPR and FNR)}
		\label{fig:FPR_FNR_Crime}
	\end{subfigure}
	\newline
	\begin{subfigure}{0.99\columnwidth}
		\centering	
		\includegraphics[scale=0.55]{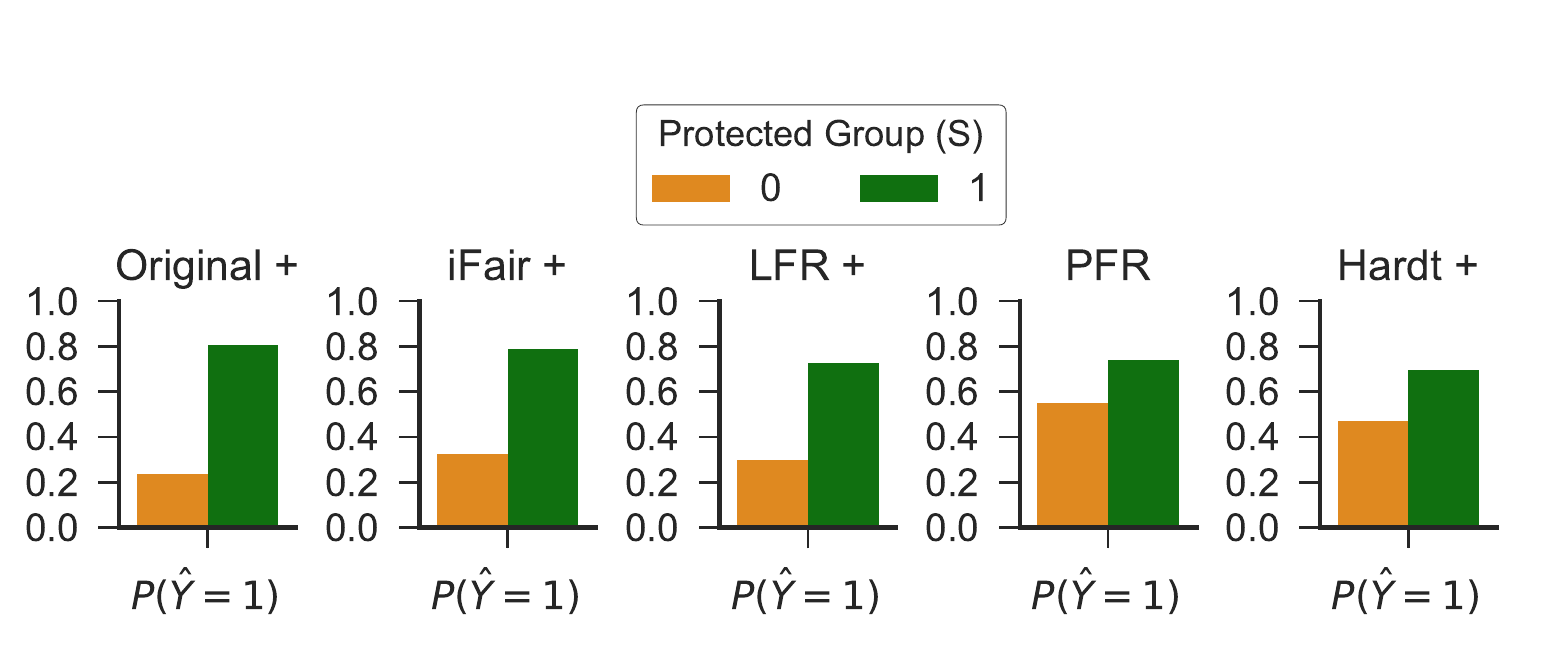}
\vspace{-0.3cm}
		\caption{Difference in per-group rates of positive prediction}
		\label{fig:Parity_Crime}
	\end{subfigure}
\vspace{-0.1cm}
	\caption{Crime \& Communities data: (a) error rates and (b) positive prediction rates.}
	\label{fig:GroupFairness_Crime}
\end{figure}

\spara{[Q4] Influence on Group Fairness}
Figure \ref{fig:FPR_FNR_Crime} shows the per-group error rates, and \ref{fig:Parity_Crime} shows the per-group positive prediction rates. 
Smaller differences in the values between the two groups are preferable.
The following observations are notable:
\squishlist
\item Disparate Mistreatment (aka. Equality of Odds): 
\emph{PFR} significantly outperforms all other methods on balancing the
error rates of  the two groups. 
Furthermore, it achieves nearly equal error rates comparable to 
the \emph{Hardt+} model, whose sole goal is to achieve equal error rates
between groups via post-processing.
\item Disparate Impact (aka. Demographic parity): 
\emph{PFR} clearly outperforms all the methods by achieving near perfect balance 
(i.e., near-equal rates of positive predictions).
\squishend

\subsubsection{Results on Compas Dataset}
\label{subsec:compas_experiments}

The results for the Compas dataset are mostly in line with the results for 
the synthetic data (in Subsection  \ref{subsec:synthetic_experiments}) and 
Crime \& Communities datasets (in Subsection  \ref{subsec:crime_experiments}).
Therefore, we report only briefly on them.

\begin{figure}[tbh!]
	\centering	
	\includegraphics[scale=0.55]{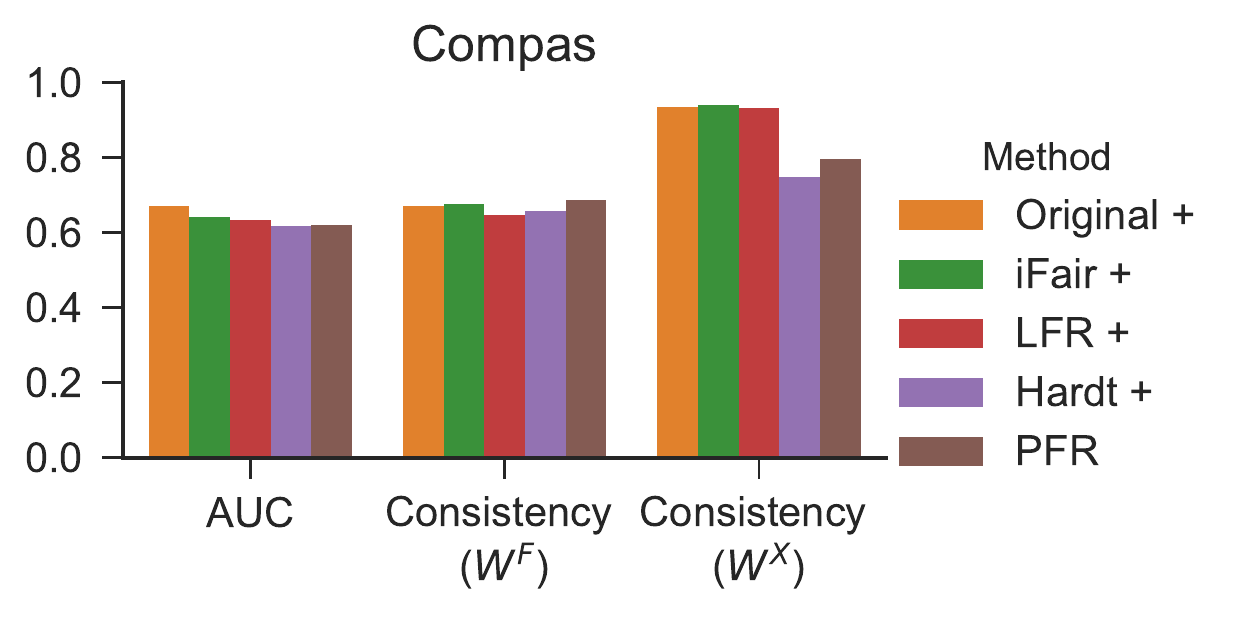}
\vspace{-0.4cm}
	\caption{Compas data: utility vs. individual fairness (higher is better).}
	\label{fig:Optim-consistency_AUC_WX_WF_Compas}
	\vspace{-4mm}
\end{figure}

\spara{\bf Utility vs. Individual Fairness:} \emph{PFR} performs similarly as the
other  representation learning methods in terms of utility and individual fairness on $W^F$, as shown in Figure \ref{fig:Optim-consistency_AUC_WX_WF_Compas}. 

\begin{figure}[tbh!]
	\begin{subfigure}{0.95\columnwidth}
		\centering	
		\includegraphics[scale=0.55]{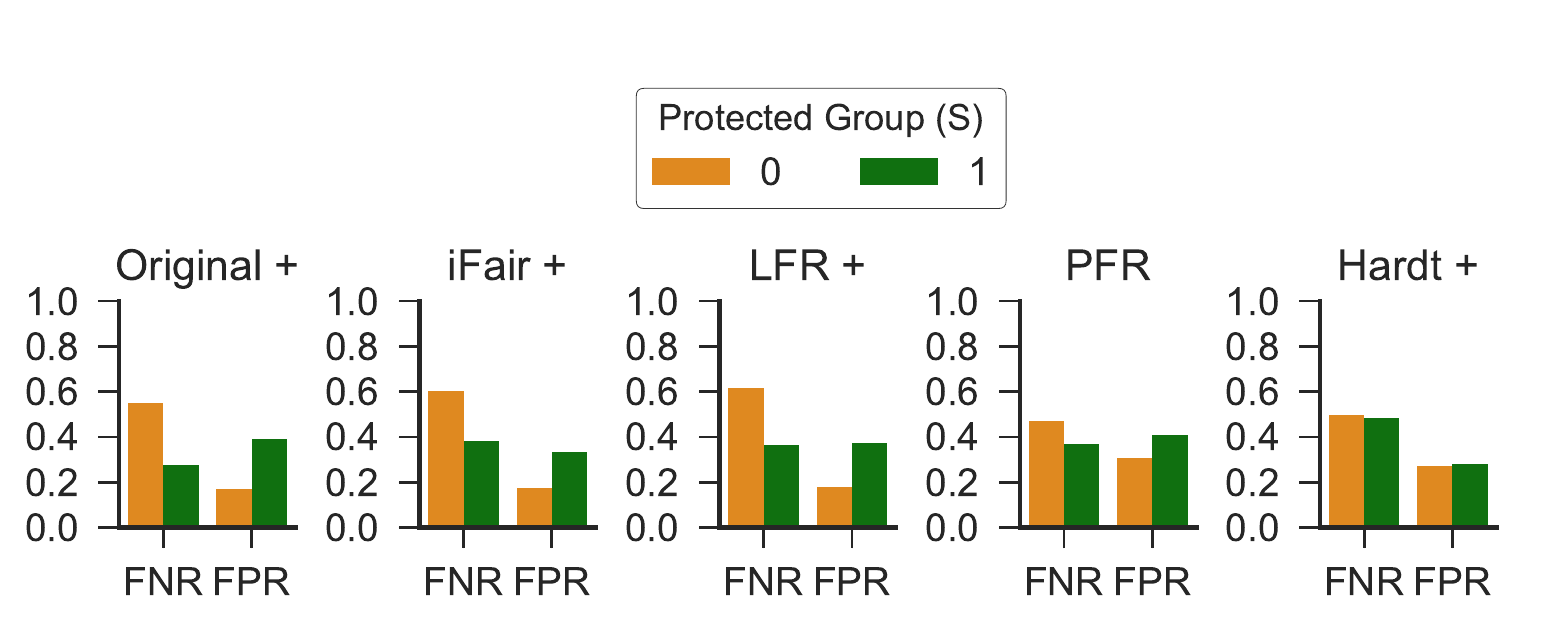}
\vspace{-0.4cm}
		\caption{Difference in per-group error rates (FPR and FNR)}
		\label{fig:FPR_FNR_Compas}
	\end{subfigure}
	\newline
	\begin{subfigure}{0.95\columnwidth}
		\centering	
		\includegraphics[scale=0.55]{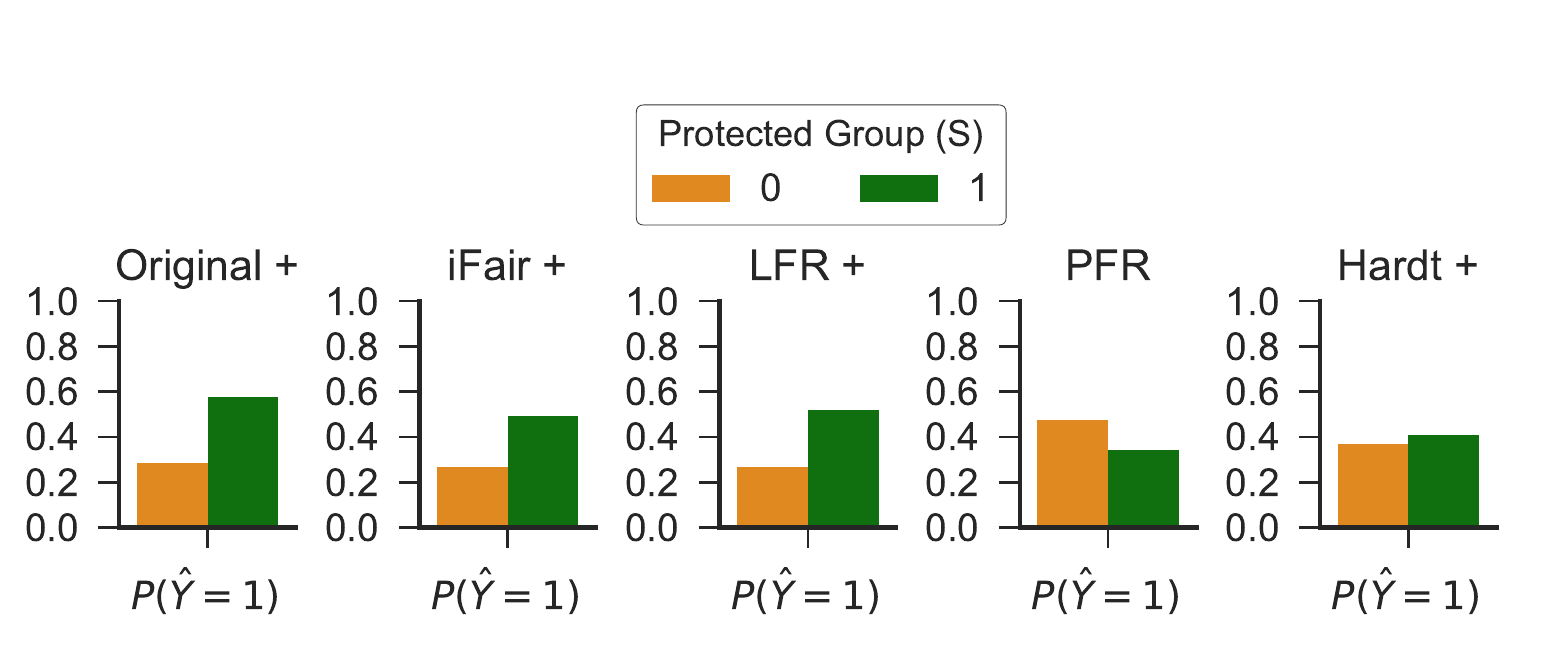}
\vspace{-0.4cm}
		\caption{Difference in per-group rates of positive prediction}
		\label{fig:Parity_Compas}
	\end{subfigure}
\vspace{-0.2cm}
	\caption{Compas data:  (a) error rates and (b) positive predictions rates.}
	\label{fig:GroupFairness_Compas}
	\vspace{-2mm}
\end{figure}

\spara{\bf Group Fairness:} However, \emph{PFR} clearly outperforms all other methods on group fairness. It achieves near-equal rates of positive predictions as shown in Figure \ref{fig:Parity_Compas}, and  near-equal error rates across groups as shown in Figure \ref{fig:FPR_FNR_Compas}. 
Again, \emph{PFR}'s performance on group fairness is as good as that of \emph{Hardt+} which is solely designed for equalizing error rates by post-processing the classifier's outcomes.

\subsection{Influence of PFR Hyper-Parameter {\bf $\gamma$}}
\label{subsec:influence_of_gamma}
{
In this subsection we analyze the influence of $\gamma$ on the trade-off between individual fairness (consistency $W^F$) and utility (AUC) of the downstream classifiers. To this end, we keep all other hyper-parameters
set to their values for the best result in the main experiments, and systematically vary the value of hyper-parameter $\gamma$ in [0,1]. 

Recall that {\em PFR} aims to preserve local neighborhoods in the input space $X$ (given by $W^X$), as well as the similarity 
given by the fairness graph $W^F$, where the hyper-parameter $\gamma$ controls the relative influence of $W^X$ and $W^F$. 
Figure \ref{fig:influence_of_gamma} shows the influence of $\gamma$ on individual fairness and utility for (a) low-dimensional synthetic, (b) Crime and (c) Compas data, respectively. We make the following key observations.

\spara{Individual Fairness:} We observe that 
with increasing $\gamma$ the consistency with regard to $W^F$ increases. This is in line with our expectation: as  $\gamma$ increases the influence of  $W^F$ on the objective function, the performance of the model on individual fairness (consistency $W^F$) improves. 
This trend holds for all the datasets.
It is worth highlighting that the improvement in individual fairness is
for newly seen test samples that were unknown at the time
when 
the fairness graph $W^F$ was constructed
and the \emph{PFR} model was learned.
This demonstrates the ability of \emph{PFR} to generalize individual fairness to unseen data.

\spara{Utility:} The influence of $\gamma$ on the utility is more nuanced. %
We observe that the extent of the trade-off between individual fairness in $W^F$ and utility 
depends on the degree of conflict between the pairwise $W^F$, and the classifier's ground-truth labels.
\squishlist
\item If $W^F$ indicates equal deservingness for data points that have different ground-truth labels, there is a natural conflict between individual fairness and utility. 
We observe this case for the real-world datasets Crime and Compas where $W^F$ is in tension with ground-truth labels -- presumably due to implicit anti-subordination embedded in graph
or equivalently, due to historic discrimination in the classification ground-truth. 
With increasing $\gamma$, there is a slight drop in the utility \emph{AUC} for the non-protected group.
However, there is an improvement in \emph{AUC}  for the protected group.
The overall AUC drops by a few percentage points, but stays at a
high level even for very high $\gamma$.
So we trade off a substantial gain in individual fairness for
a small loss in utility.
This is a clear case of how incorporating side-information on pairwise judgments can help to improve algorithmic decision making for
historically disadvantaged 
groups.

\item In contrast, if $W^F$ pairs of equal deservingness 
are compatible with the classifier's ground-truth labels, there is no trade-off between utility and individual fairness. 
In such cases, $W^F$ may even help to improve the utility 
by better learning a similarity manifold in the input space.
We observe this case for the synthetic data where $W^F$ 
is consistent with the ground-truth labels. As $\gamma$ increases, the AUC of 
a classifier trained on \emph{PFR} is enhanced. The improvement in \emph{AUC} holds
for both protected and non-protected groups.

\squishend

\begin{figure}[t!]
	\centering	
	\begin{subfigure}{0.45\columnwidth}
		\centering	
		\includegraphics[scale=0.42]{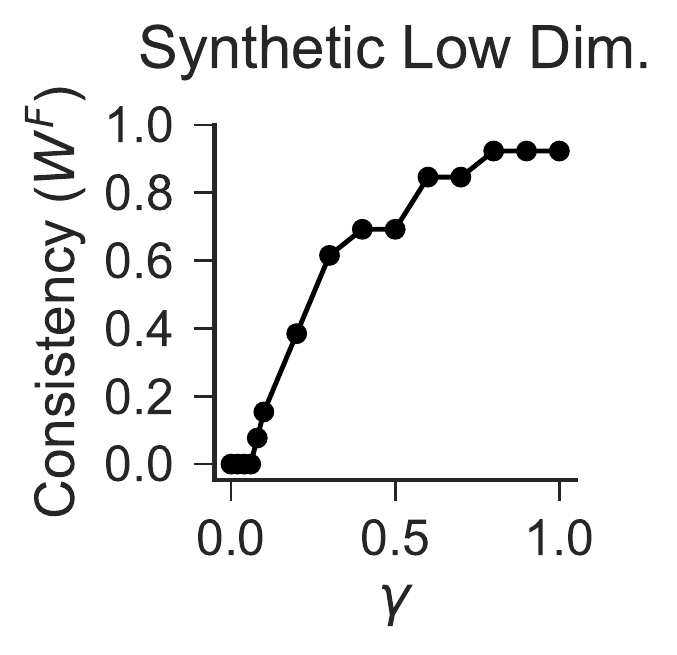}			
	\end{subfigure}
	\begin{subfigure}{0.53\columnwidth}
		\centering	
		\includegraphics[scale=0.40]{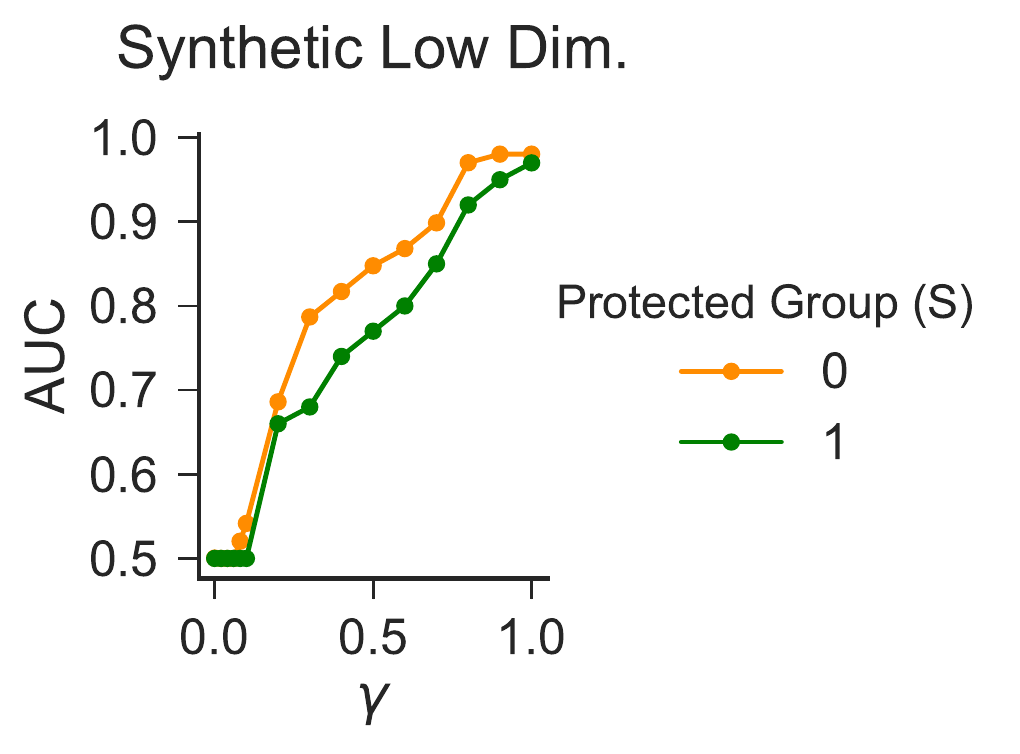}
	\end{subfigure}
	\newline
	\begin{subfigure}{0.45\columnwidth}
		\centering	
		\includegraphics[scale=0.38]{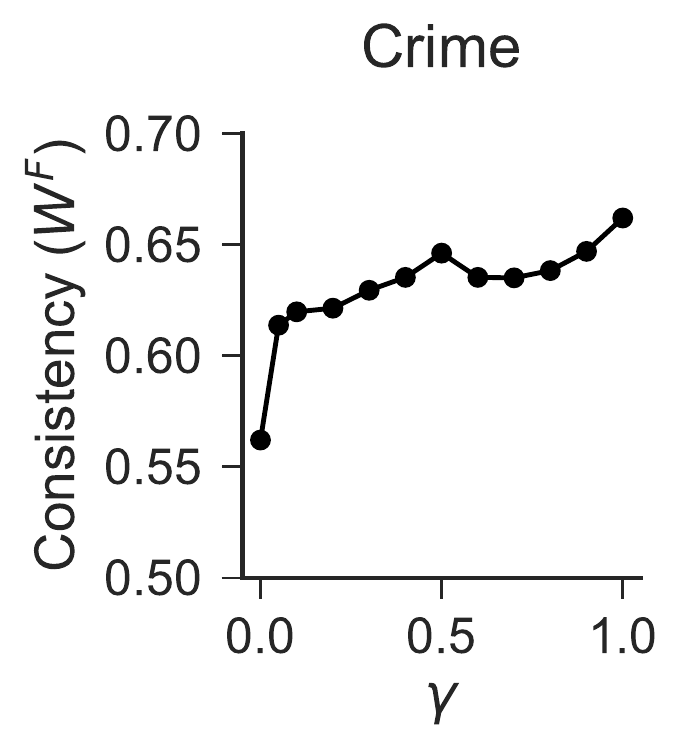}			
	\end{subfigure}
	\begin{subfigure}{0.53\columnwidth}
		\centering	
		\includegraphics[scale=0.39]{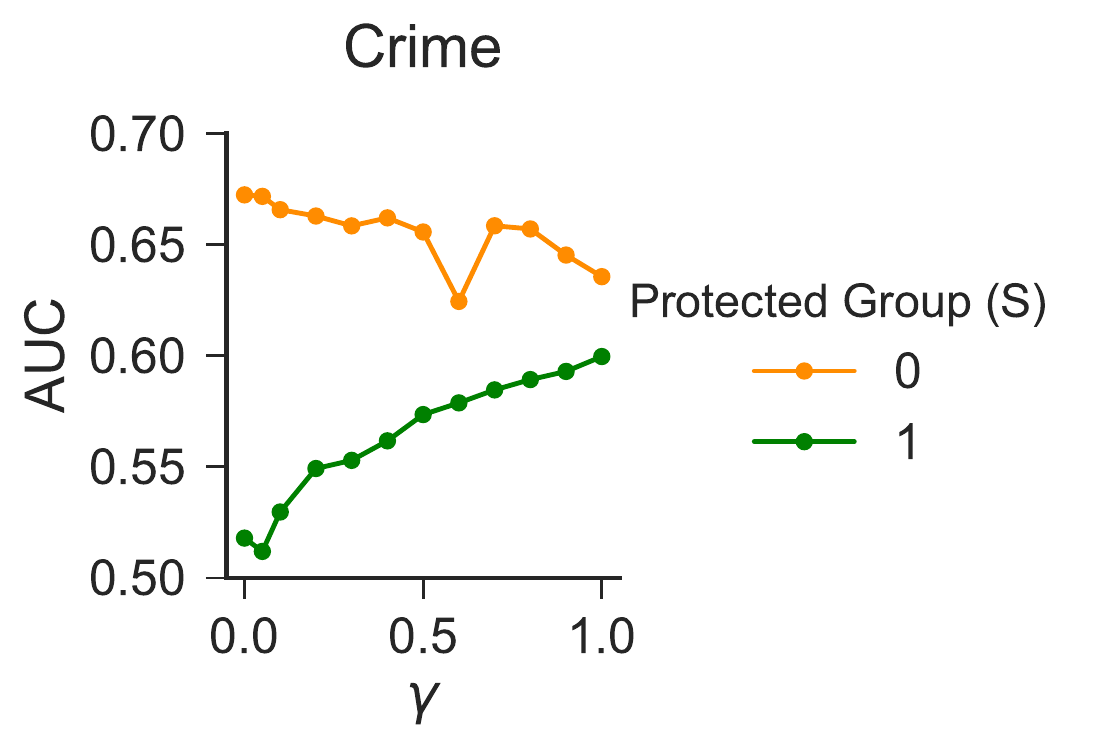}
	\end{subfigure}
	\newline
	\begin{subfigure}{0.45\columnwidth}
		\centering	
		\includegraphics[scale=0.38]{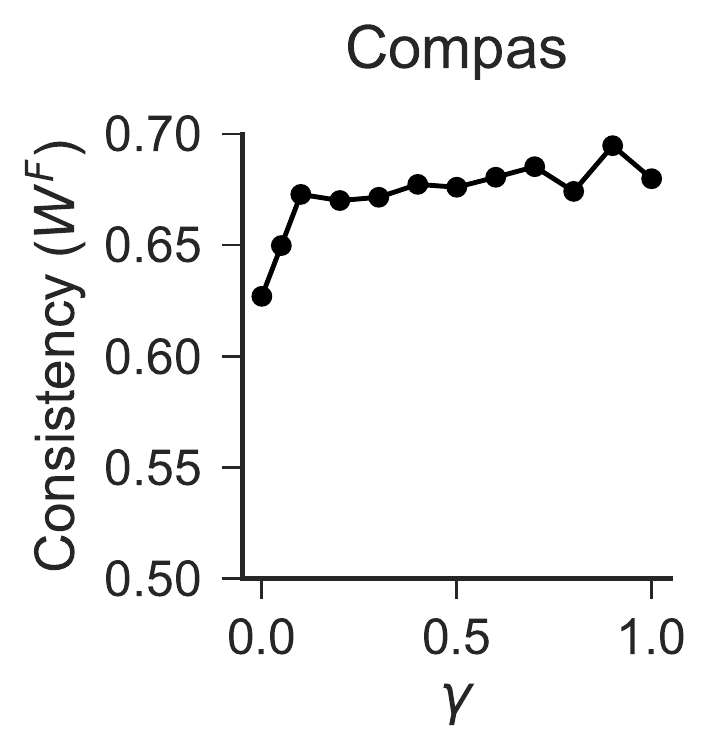}
	\end{subfigure}
	\begin{subfigure}{0.53\columnwidth}
		\centering	
		\includegraphics[scale=0.39]{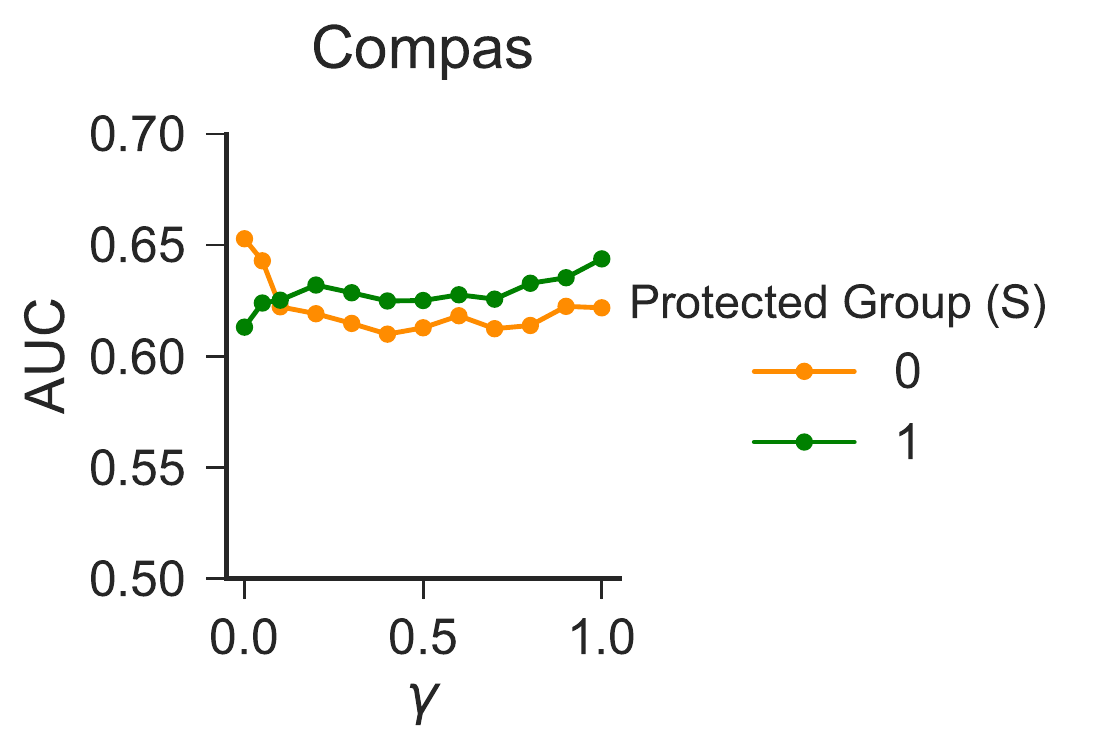}
	\end{subfigure}
\vspace{-0.3cm}
	\caption{Influence of $\gamma$ on 
individual fairness
		and utility.}
	\label{fig:influence_of_gamma}
	\vspace{-2mm}
\end{figure}
}

\subsection{Sensitivity to Sparseness of Fairness Graph}
\label{subsec:sensitivity_to_labelsize}
{
In this section we study the sensitivity of 
 \emph{PFR} to the sparseness of the labeled pairs in the
fairness graph $W^F$.
We fix all hyper-parameters to their best values in the
main experiments, and systematically vary the fraction of 
datapoints for which we use pairwise fairness labels.
 The results are shown in Figure \ref{fig:influence_of_labelsize}. 
All results reported are on out-of-sample withheld test set of fairness graph $W^F$. 
Recall that \emph{PFR} accesses fairness labels only for training data. 
For test data, it solely has the data attributes available in $X$.

\spara{Setup:} For the synthetic data, 
we uniformly at random sampled fractions of 
[$log_2\,N$, $\frac{N}{5}$, $\cdots$, $N$, $N\,log_2\,N$, $N^2$] pairs from the training data, which for this data translates into 
$[9, 120, \cdots 600,  5537, 360000]$ pairs. 
For the Crime data, 
we varied the percentage of training samples
for which use equivalence labels, in steps of 10\% from 10\% to 100\%.
For the Compas data, 
we varied the percentage of training data points for which we elicit 
per-group rankings, in steps of 10\% from 10\% to 100\%.

\spara{Results:} We observe the following trends.
\squishlist
\item Increasing the fraction of fairness labels improve
the results on individual fairness (consistency for $W^F$),
while hurting utility (\emph{AUC}) only mildly (or even improving it in certain cases).
\item For the synthetic data, 
even with as little as 0.17\% of the fairness labels, the results
are already fairly close to the best possible:
consistency for $W^F$ is already 90\%,
and \emph{AUC} reaches 95\%.
\item For the Crime data, 
we need about 30 to 40\% to get close to the best results for
the full fairness graph.
However, even with sparseness as low as 10\%, 
\emph{PFR} degrades smoothly:  consistency $W^F$ is 59\% compared to 68\% for the full graph, and \emph{AUC} is affected only mildly by
the sparseness.
\item For the Compas data, 
we observe similar trends: even with very sparse
$W^F$ we stay within a few percent of the best possible
consistency, and \emph{AUC} varies only mildly with changing sparseness
of the fairness graph.

\squishend
\begin{figure}[t!]
	\centering	
	\begin{subfigure}{0.32\columnwidth}
		\centering	
		\includegraphics[scale=0.31]{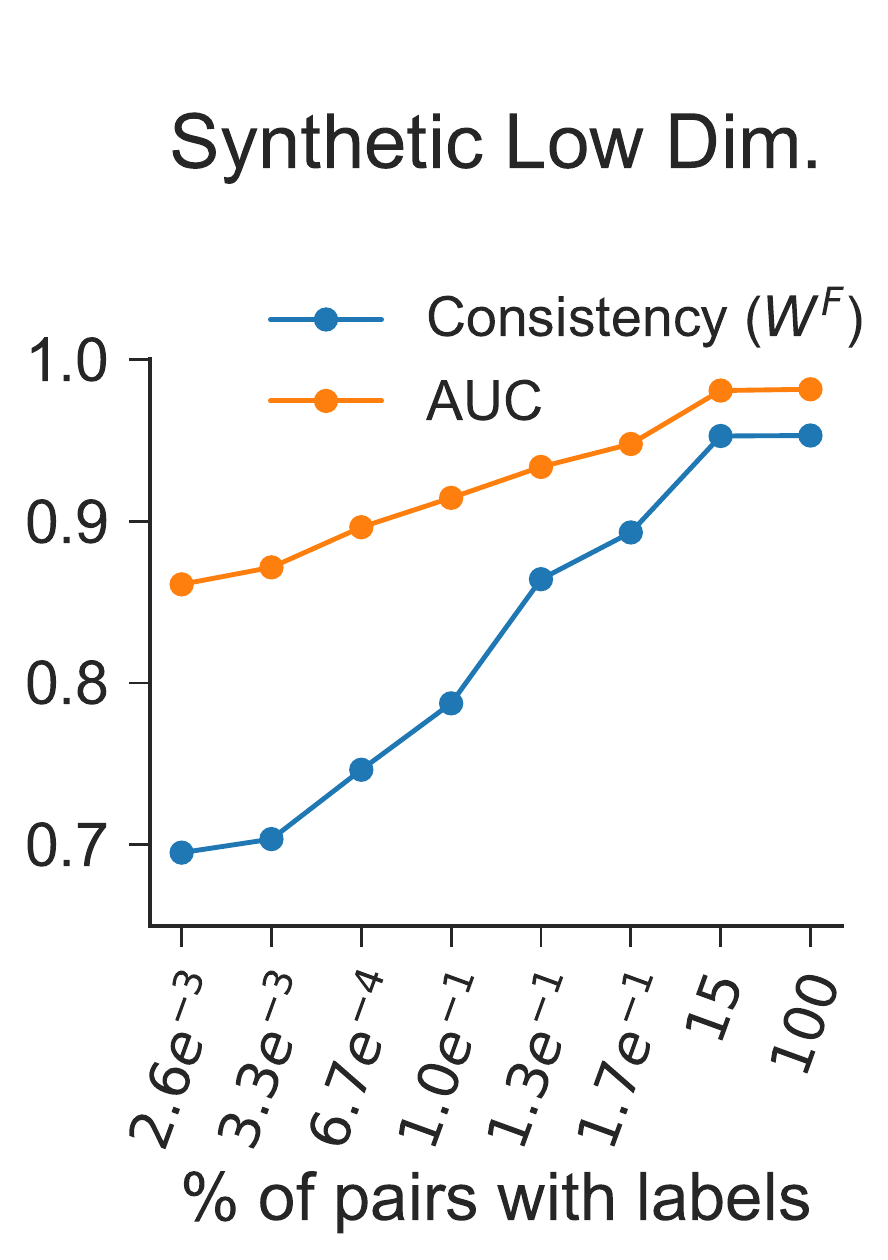}
	\end{subfigure}
	\begin{subfigure}{0.32\columnwidth}
		\centering	
		\includegraphics[scale=0.31]{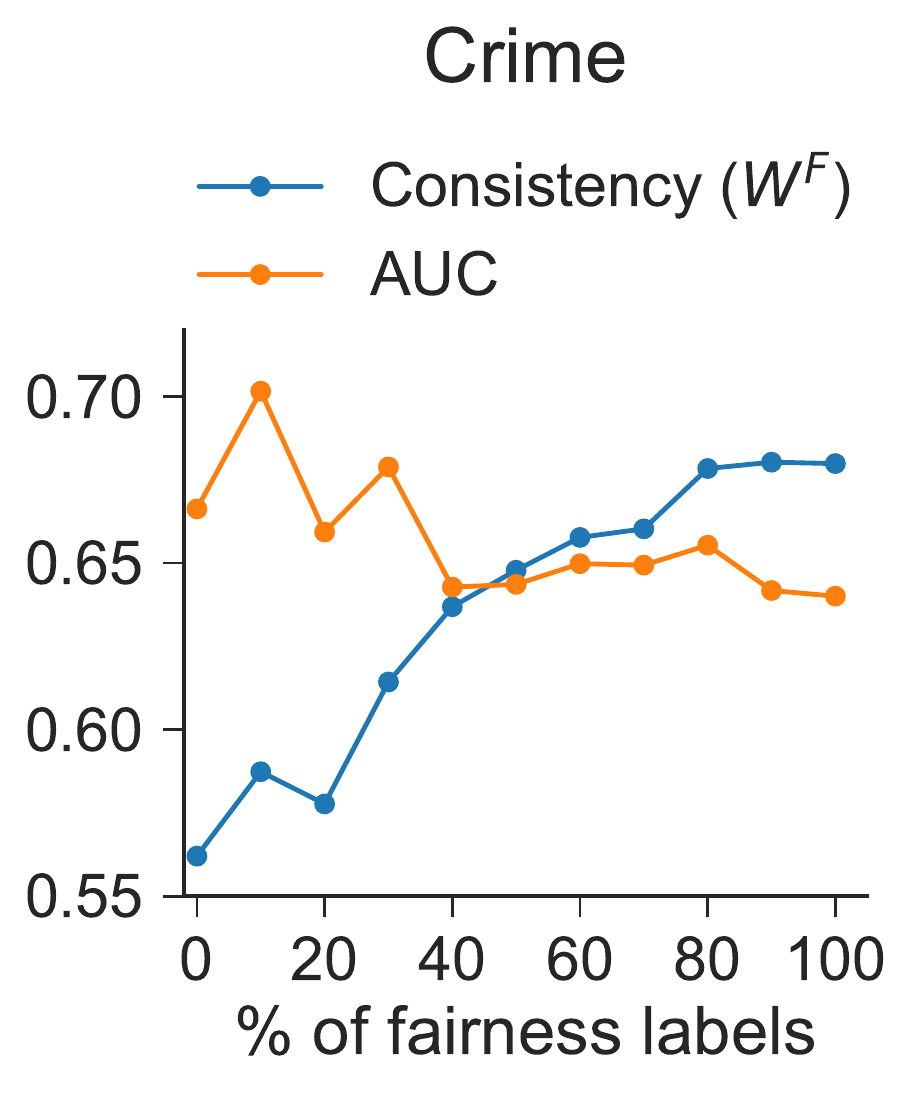}			
	\end{subfigure}
	\begin{subfigure}{0.32\columnwidth}
		\centering	

		\includegraphics[scale=0.31]{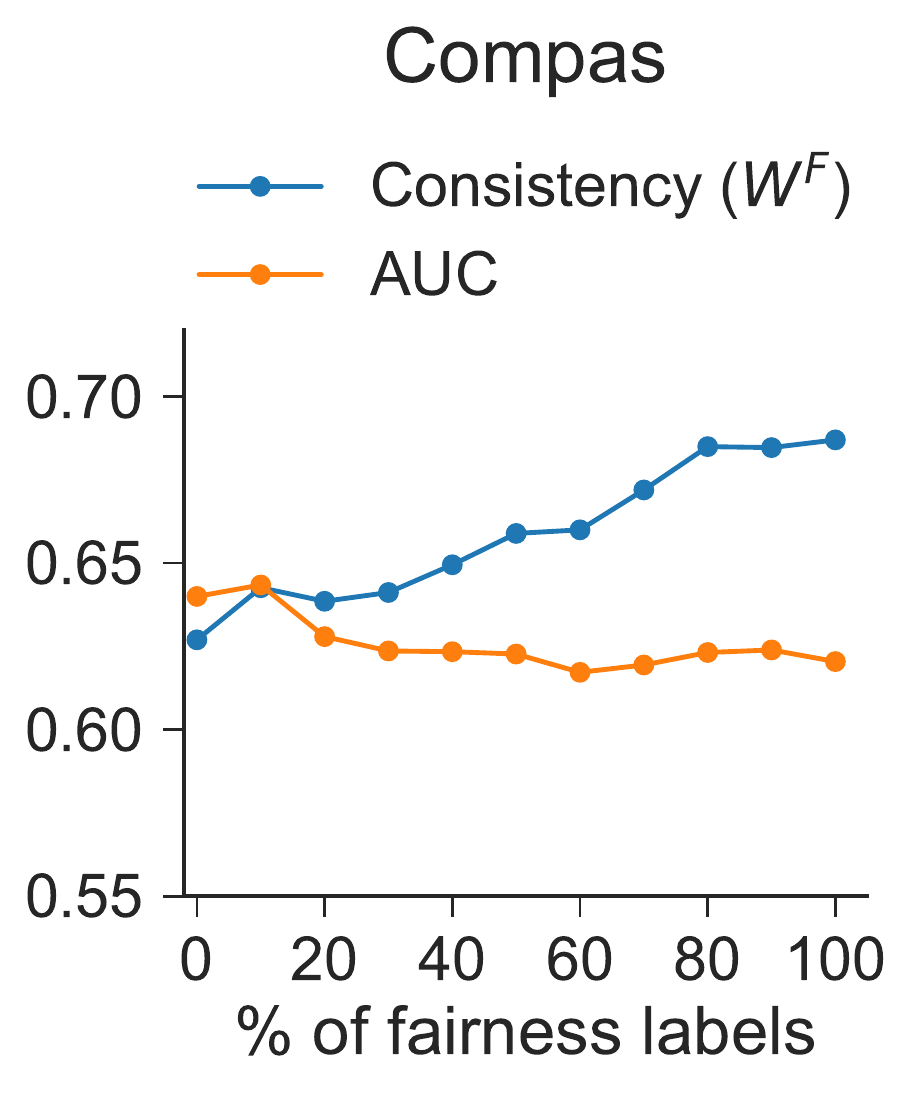}		
	\end{subfigure}
\vspace{-0.2cm}
	\caption{Influence of 
fairness-graph sparseness. 
}
	\label{fig:influence_of_labelsize}
	\vspace{-2mm}
\end{figure}

These observations indicate that the \emph{PFR} model yields benefits already
with a small amount of human judgements of equally deserving 
individuals.
}

\balance

\subsection{Discussion and Lessons}
{

%

\noindent The experimental results suggest several
key findings.

\squishlist
\item {\em Individual Fairness - Utility Trade-off:}
The extent of 
this trade-off
depends on the degree of conflict between the 
fairness graph and the classifier's ground-truth labels. 
When edges in the fairness graph connect data points (for equally deserving individuals)
that have different ground-truth labels, there is an inherent tension between individual fairness and utility. 
For datasets where
some compromise is unavoidable, \emph{PFR}
turns out to perform best in balancing the different goals.
It is consistently best with regard to individual fairness,
by a substantial margin over the other methods.
On utility, its \emph{AUC} is competitive and always close to the
best performing method on this metric, typically within
2 percentage points of the best \emph{AUC} result.
\item {\em Balancing Individual Fairness and Group Fairness:}
The human judgements cast into the fairness graph
help \emph{PFR} to perform well also on group fairness criteria.
On these measures, \emph{PFR} is almost as good as the method
by Hardt et al., which is specifically geared for group fairness
(but disregards individual fairness).
To a large extent, this is because the pairwise fairness 
judgments address historical subordination of groups.
Eliciting human judgements is a crucial asset for fair
machine learning in a wider sense.
\item {\em Data Representation:}
The graph-embedding method used by \emph{PFR} appears
to the best way of incorporating the pairwise human judgements.
Alternative representations of the same raw information such as additional features in the input dataset, as leveraged by the augmented baselines
(\emph{LFR+}, \emph{iFair+}), perform considerably worse than
\emph{PFR} on consistency ($W^F$).

The $W^F$ input is needed solely for the training data;
previously unseen test data (at deployment of the learned
representation and downstream classifier) does not have any
pairwise judgments at all. This underlines the practical viability of \emph{PFR}.

\item {\em Graph Sparseness:} 
Even a small amount of pairwise fairness
judgments helps \emph{PFR} in improving fairness.
At some point of extreme sparseness, \emph{PFR} loses this advantage,
but its performance degrades quite gracefully. 
\item {\em Robustness:} 
PFR is fairly robust to the dimensionality of the dataset. As the dimensionality of the input data increases, the performance
of \emph{PFR} drops a bit, but still outperforms other approaches
in terms of balancing fairness and utility.
Furthermore, 
\emph{PFR} is quite insensitive to the choice of hyper-parameters. Its performance remains
stable across a  wide range of values.
\item {\em Limitations:}
When the data exhibits a strong conflict between fairness and
utility goals, even \emph{PFR} will fail to counter such tension
and will have to prioritize either one of the two criteria
while degrading on the other.
The human judgements serve to mitigate exactly such
cases of historical subordination and discrimination, 
but if they are too sparse or too noisy, their influence
will be marginal. 
For the datasets in our experiments, we assumed that the information
on equally deserving indidivuals would reflect high consensus 
among human judges. When this assumption is invalid for certain
datasets, \emph{PFR} will lose its advantages
and perform as poorly as (but no worse than) other methods.
\squishend
} %
\section{Conclusions}

This paper proposes a new departure for the hot topic of
how to incorporate fairness in algorithmic decision making.
Building on the paradigm of individual fairness, 
we devised a new method, called \emph{PFR}, for operationalizing this line of
models, by eliciting and leveraging 
side-information on pairs of individuals who are equally deserving and, thus, 
should be treated similarly for a given task.
We developed a representation learning model to learn
Pairwise Fair Representations (\emph{PFR}), 
as a fairness-enhanced input to downstream machine-learning tasks.
Comprehensive experiments, with synthetic and real-life datasets, indicate that 
the pairwise judgements
are
beneficial for 
members
of the protected group,
resulting in high individual fairness and
high group fairness (near-equal error rates across groups)
 with reasonably low loss in utility.

 \section{Acknowledgment} This research was supported by the ERC Synergy Grant ``imPACT'' (No. 610150) and ERC Advanced Grant ``Foundations for Fair Social Computing'' (No. 789373).

\balance
\bibliographystyle{abbrv}
\bibliography{references}
\end{document}